\documentclass[pdflatex,sn-basic]{sn-jnl}

\usepackage{graphicx}%
\usepackage{multirow}%
\usepackage{amsmath,amssymb,amsfonts}%
\usepackage{amsthm}%
\usepackage{mathrsfs}%
\usepackage[title]{appendix}%
\usepackage{xcolor}%
\usepackage{textcomp}%
\usepackage{manyfoot}%
\usepackage{booktabs}%
\usepackage{caption}
\usepackage{algorithm}%
\usepackage{algorithmicx}%
\usepackage{algpseudocode}%
\usepackage{listings}%
\usepackage{tabularx}%
\usepackage{booktabs}%
\usepackage{graphicx}%
\usepackage[table]{xcolor}%

\newcommand{\redtext}[1]{\textcolor{black}{#1}}
\newcommand{\bluetext}[1]{\textcolor{black}{#1}}

\usepackage{tikz}%
\usepackage{pgfplots}
\pgfplotsset{compat=1.18}
\usetikzlibrary{positioning, arrows.meta} 
\usepackage{enumitem}
\usepackage{tabularx,booktabs,array,multirow,graphicx}
\usepackage[table]{xcolor}
\newcommand{\rot}[1]{\rotatebox{90}{\scriptsize\strut #1}}

\usepackage{pifont}
\usepackage{xcolor}

\newcommand{\cmark}{\textcolor{green!60!black}{\ding{51}}}
\newcommand{\xmark}{\textcolor{red!70!black}{\ding{55}}}
\newcommand{\lmark}{\textcolor{orange!85!black}{\ding{115}}}

\usepackage{xltabular}

\usepackage[most]{tcolorbox}

\theoremstyle{thmstyleone}%
\theoremstyle{thmstyletwo}%

\theoremstyle{thmstylethree}%

\begin{document}

\title[Narratology LLM Survey]{Narrative Theory-Driven LLM Methods for Automatic Story Generation and Understanding: A Survey}


\author*[1]{\fnm{David Y.} \sur{Liu}}\email{david.liu2@unsw.edu.com.au}
\author[1]{\fnm{Aditya} \sur{Joshi}}\email{aditya.joshi@unsw.edu.au}
\author[2]{\fnm{Paul} \sur{Dawson}} \email{paul.dawson@unsw.edu.au}

\affil[1]{\orgdiv{School of Computer Science and Engineering}}
\affil[2]{\orgdiv{School of Arts and Media}}
\affil{\orgname{University of New South Wales (UNSW)}, \orgaddress{\city{Sydney}, \postcode{2052}, \state{NSW}, \country{Australia}}}


\abstract{
     Applications of narrative theories using large language models (LLMs) deliver promising methods in automatic story generation and understanding tasks. Our survey examines how natural language processing (NLP) research uses LLM methods to engage with diverse concepts from narrative studies. We use established distinctions from narratology to categorise ongoing efforts and discover the following: \redtext{(a) narrative texts come from diverse sources beyond just literature, (b) theoretical synthesis and validation are potential outcomes, (c) generation tasks lag behind understanding in several ways: theoretical application, post-training methods, exploring non-fiction narratives and addressing narrative levels beyond fabula and discourse.} For future directions, instead of the pursuit of a single, generalised benchmark for `narrative quality', we believe that progress can benefit from efforts that focus on the following: defining and improving theory-based metrics for individual narrative attributes; continue conducting large-scale, theory-driven literary/social/cultural analysis; generating narratives in situated contexts; and continuing experiments where outputs can be used to validate or refine narrative theories. This work provides a contextual foundation for more systematic and theoretically informed narrative research in NLP by providing an overview to ongoing research efforts and the broader narrative studies landscape.



}


\maketitle

\section{Introduction}\label{sec1}

    Storytelling is predominantly associated with literature, but is also a widespread communicative practice across domains such as journalism, social media discourse and everyday conversation. Narrative is understood as both a textual artefact and a mode of human cognition \citep{bruner1991narrative}. For example, the term `narrative' can refer to both a news article text reporting on climate change, and the cognitive modes of ``act now to avert disaster'' versus ``already too late to make a difference.'' \redtext{Extending this cognitive perspective, we also observe an increasing support for the narrative-first hypothesis of human language \redtext{which questions the assumption of narrative as only a feature of language, and proposes the idea that} spoken and written languages emerged primarily as tools for storytelling \citep{turner1996literary,mcbride2014storytelling,ferretti_narrative_language_2024}.} 
    
    Following this, it is unsurprising to see that story understanding and generation form key tasks in Natural Language Processing (NLP). For example, automatic story understanding systems are useful for ensuring online safety by detecting information intended to manipulate public opinion, or gathering sequential information from multiple fragmented news sources and forums \citep{ranade2022computational}; automatic story generation (ASG) systems have potential applications in gaming, education, mental health, and marketing \citep{chhun2024language}. \redtext{In their suggestions to progress the field,} \cite{alhussain2021automatic} encourage ASG researchers to integrate literary knowledge and narrative theories, believing such integration enhances coherence, creativity, emotional engagement, and alignment with human storytelling conventions.
    
    We find that examples of theoretical integration mostly draw from narratology, \redtext{the systematic study of narrative}, which originated in literary theory to examine the logic, principles, and practices of narrative representation \citep{Meister2014Narratology}. There also exist narrative theories outside of narratology, where broader narrative studies extend across the humanities and social sciences.\footnote{Since the `narrative turn' that launched in the 1980s, the academic study of narrative as a fundamental form of human consciousness has developed alongside and beyond literary studies in multiple disciplines, such as law, psychology, social sciences and economics. There is no unified knowledge about narrative shared between the disciplines, the fields have differing methodologies and objects of study. To learn more about this development from the perspective of narratology, see \citep{kreiswirth1992trusting,hyvarinen2006towards,dawson2017many}.} \redtext{This survey situates ongoing efforts within this broader interdisciplinary landscape to better recognise the inspirations, assumptions and limitations shaping recent NLP methods for story generation and understanding}.
    
    Current NLP methods have integrated recent advancements in Large Language Models (LLMs)\footnote{\redtext{By LLMs, we specifically mean transformer-based, pre-trained, decoder-only language models with more than 1 billion parameters.}}, where their advantage in generalisation over unseen tasks makes it possible to apply narrative theories through prompting to extract, classify, and generate textswith reduced dependency on massive and costly annotated datasets \citep{marino2024integrating, piper-bagga-2024-using}. LLMs offer low-cost and practical solutions to conduct theory-driven narrative analysis at scale \citep{carroll2024towards}, improving the accessibility of interdisciplinary collaboration \redtext{across fields such as Computational Humanities Research (CHR) and Digital Humanities (DH) \citep{Mozaffari_literary_llm_survey}. }

\begin{figure}[htbp]
        \centering
        \includegraphics[width=0.8\linewidth]{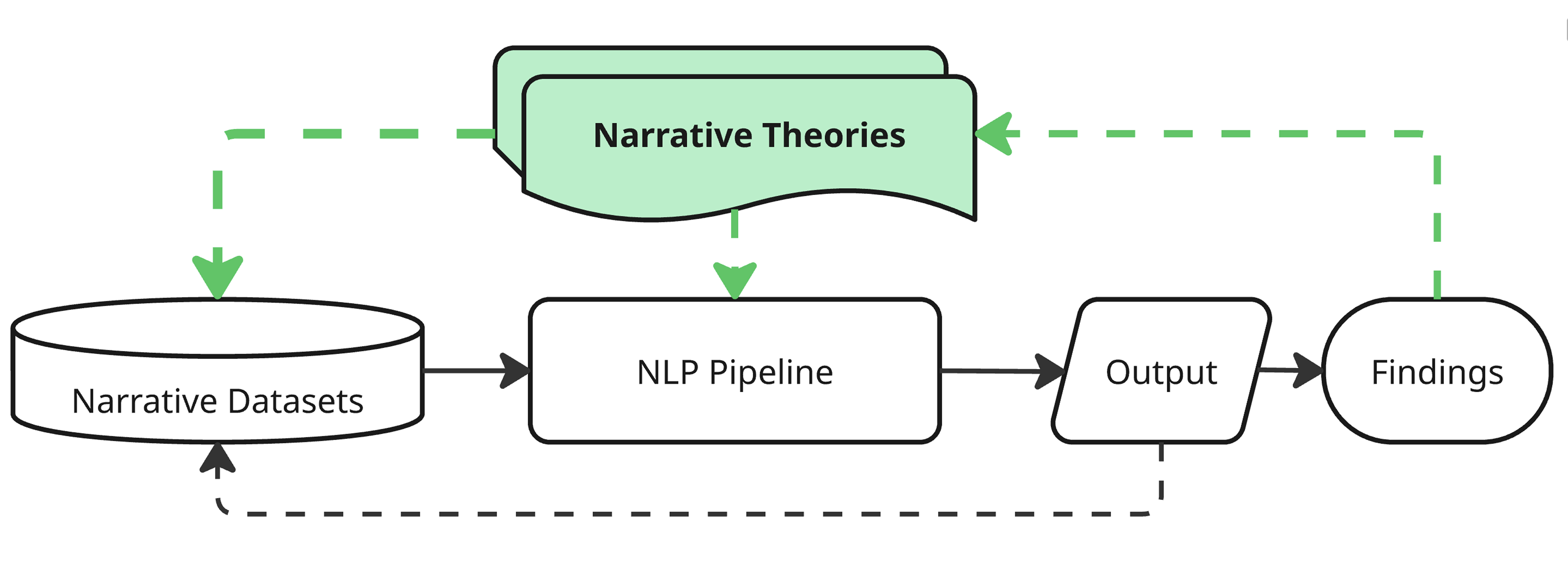} 
        \caption{Illustration of the bi-directionality between narrative theories and NLP.}
        \label{fig:narrative_bidirectional}
    \end{figure}

    Despite \bluetext{the growing interest in LLM methods for interdisciplinary narrative studies}, existing surveys do not systematically examine how narrative theories influence the design and usage of LLM methods (Section \ref{sec_related_work}). \redtext{Following existing efforts of using narrative theories as organisational frameworks for ongoing research \citep{piper2021narrative,piper-2023-computational,gervas2024storytelling,hamilton-etal-2026-narrabench},} our survey highlights how theoretical narrative frameworks inform computational pipelines and, conversely, how the research of new LLM methods can offer insights to potentially refine, synthesise, and validate those very theories. In this sense, NLP methods serve both as consumers of narrative theories and potential contributors (Figure~\ref{fig:narrative_bidirectional}), where the results of computational analysis can in turn be used to formulate insights relating to the applied theories. \redtext{This bi-directionality echoes similarly identified processes in digital literary studies \citep{hatzel2023machine}, humanities and social sciences \citep{pichler2020reflektierte}. Therefore, a survey of narrative theory-driven LLM methods is a necessary step toward bridging computational practice and the broader theoretical traditions of narrative studies.} 

    Our survey aims to clarify the connections in Figure \ref{fig:narrative_bidirectional} by answering the questions:
    
    \begin{itemize}
        \item[] \textbf{(1) What narrative theories are commonly used by LLM methods, and what textual datasets and tasks do LLM methods target?} We survey recent LLM-based narrative NLP methods and provide a mapping between narrative levels, datasets, tasks, and LLM methods. 
        \item[] \textbf{(2) In what ways are LLM methods used in narrative theory-informed story generation and understanding tasks, and how do they perform?} We detail the in-context, training and compound systems used by LLM methods.
        \item[] \textbf{(3) How might LLM methods contribute to the landscape of narrative studies, narrative theory and narratology?} We discuss how LLM-based methods may support theory synthesis, taxonomy design and empirical validation through large-scale narrative analysis.
    \end{itemize}


    \redtext{Our paper's contributions are organised as follows: Section \ref{sec_narrative_landscape} provides an overview of the landscape of narrative studies and introduce key narratological terminologies that we use for categorisation. Section \ref{sec_related_work} situates our work among related surveys and identifies the under-explored intersection of narrative theories and LLM methods. Section \ref{sec_scope_trends} shows the growing adoption of LLM methods over earlier computational approaches, finding that theoretically informed methods are more represented in narrative understanding than generation. Section \ref{sec_data_task} shows that a diversity of literary and non-literary textual datasets and tasks being explored by current research. Theories from literary narratology are increasingly being applied to texts outside of their original intentions. Section \ref{sec_NLP_pipeline} shows that applications of narrative theories directly shapes the NLP pipeline in labelling, modelling and evaluation. Section \ref{sec_LLM_techniques} shows that story generation tasks almost exclusively use prompt-engineering techniques while fine-tuning is only applied for understanding tasks. There is a lack of established supervised or reinforcement learning method for story generation. Section \ref{sec_outcomes} summarises five different types of research outcomes from the surveyed papers: (i) methodological and performance gains; (ii) new data, annotations, and benchmarks; (iii) insights into LLM capabilities, biases, and cultural framing; (iv) domain-specific narrative findings; and (v) theoretical and conceptual developments. Section \ref{sec_future} discusses challenges and recommends future directions in continual engagement with narrative theories beyond narratology, focus on theoretical validation/refutation as an outcome, development of specific metrics instead of overall story quality, investigate the narrative impact of pre-training and develop post-training for story generation. Finally Section \ref{sec_conclusion} concludes the survey.}

\section{\redtext{Narrative Studies Landscape and Terminology}}\label{sec_narrative_landscape}

    \begin{figure}[htbp]
        \centering
        \includegraphics[width=0.8\linewidth]{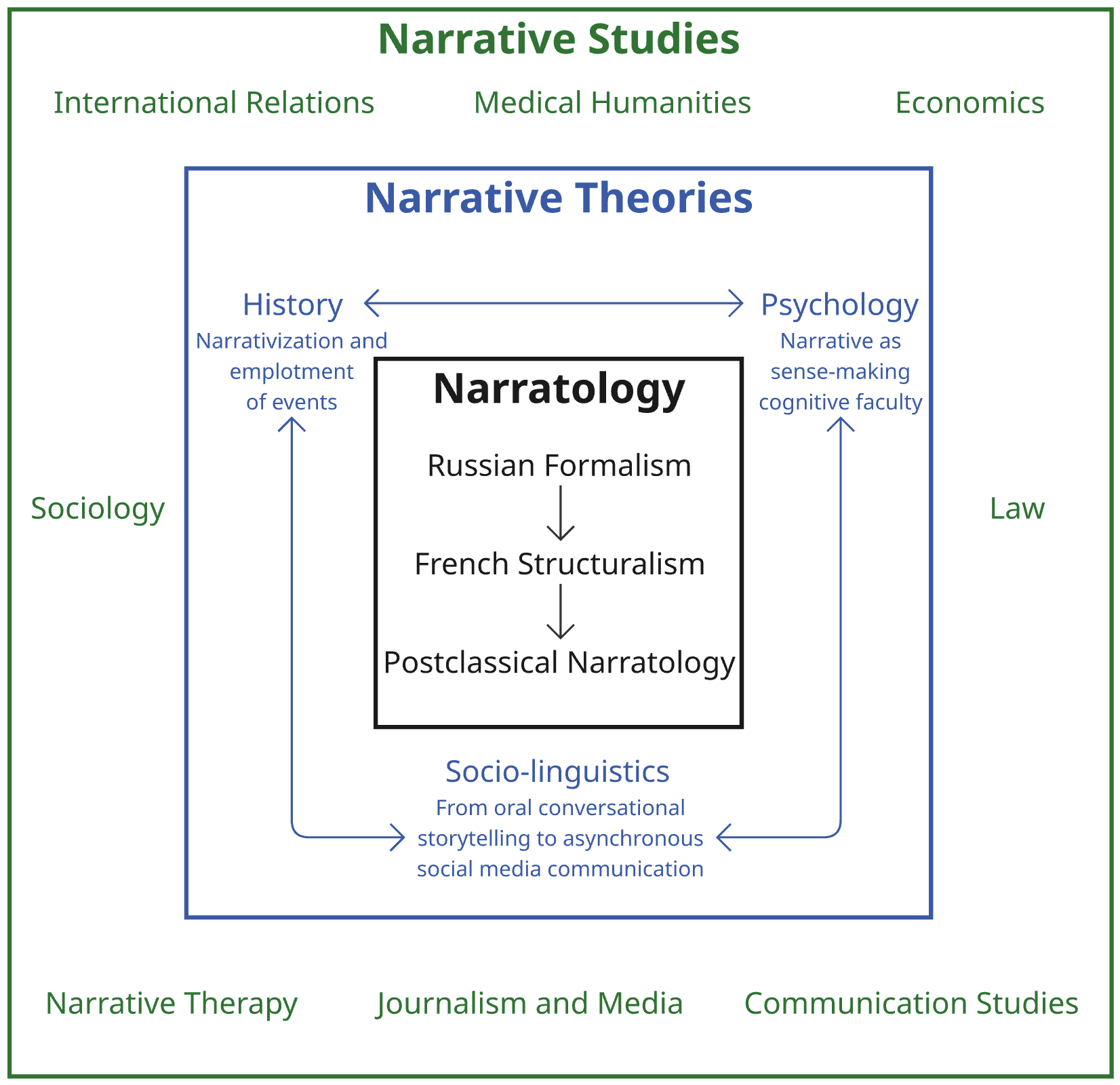} 
        \caption{Narrative Studies Landscape.}
        \label{fig:landscape}
    \end{figure}
    
   This section outlines the terminology we use to navigate the landscape of narrative studies and to position LLM-based narrative research within the existing body of interdisciplinary knowledge. It is important to first distinguish the terms \emph{narrative studies}, \emph{narrative theory}, and \emph{narratology}, as their usage can sometimes overlap (see Figure~\ref{fig:landscape}). The terminology here establishes our categorisation framework which we use in subsequent sections of the survey.
    
    \begin{description}[itemsep=0.5em, labelwidth=!, leftmargin=0pt]
        \item[\textbf{Narrative Studies:}] \redtext{We consider narrative studies as the discipline-agnostic examination of narratives as a means of shaping thought and practice.} Narratives are studied in diverse forms, including texts, cultural scripts, and strategic frameworks, across domains such as politics \citep{miskimmon2014strategic}, medicine \citep{charon2008narrative}, psychological counselling \citep{freedman1996narrative}, economics \citep{shiller2019narrative} and law \citep{brooks1996law}. Recent LLM research has also investigated narrative tasks in these domains \citep{otmakhova-frermann-2025-narrative,gupta-etal-2025-automated,kim-etal-2025-share-story,jiang-etal-2024-leveraging}. Narrative studies can explore how narratives influence meaning-making, identity, and social action, e.g., how is climate change depicted by news narratives around the world \citep{zhou-etal-2024-large,vallejo-etal-2025-human}?
    
        \item[\textbf{Narrative Theory:}] Narrative theory refers to the systematic study of how narratives function as both an object and a method of inquiry. It investigates the principles of storytelling, the construction of meaning, and the interpretive processes involved in narrative communication. Outside of literary theory, it draws from fields such as socio-linguistics \citep{de2011analyzing}, psychology \citep{narrativeidentityMcAdams} and sociology \citep{riessman2008narrative}. e.g., \cite{lincoln1999theorizing}'s \textit{Theorising Myth} asks: what role does myth play in legitimising authority, shaping collective belief, and structuring scholarship itself?
    
        \item[\textbf{Narratology:}] Narratology is the formalist study of narrative structure. It focuses on identifying the essential components of a narrative and the relationships among them, addressing the foundational question of \emph{`what makes something a narrative?'} While historically rooted in literary studies, it has since evolved to different media from films to video games, as well as phenomena beyond the text, including the contextual and cognitive dimensions of storytelling; however, its core concern with narrative structure remains unchanged \citep{sommer2012merger,ryan2014storyworlds,thon2016transmedial}. Narratology provides tools for analysing narrative form and organisation, e.g., what narrative features can be used to identify an unreliable narrator \citep{brei-etal-2025-classifying}?
    \end{description}

\subsection{Narratology}

    \redtext{We use distinction from narratology to construct our survey's taxonomy\footnote{\redtext{See also \citet{piper2021narrative,wang2023open,gervas2024storytelling,hamilton-etal-2026-narrabench}.}}, and we therefore draw on the \textit{Living Handbook of Narratology} \citep{huhn2013living} as a theoretical foundation and recommend it as a valuable reference for future research.} This section in particular draws heavily from the entries of \cite{Meister2014Narratology} and \cite{Pier2014NarrativeLevels} to briefly cover the history of narratology while introducing key terminologies that are crucial in positioning ongoing research.
    
    Emerging from formalism and structuralism, \textbf{Classical Narratology} applies structuralist insights from linguistics to the study of narrative. Formalism foregrounds intrinsic textual features such as structure, style, and linguistic form over historical or sociocultural context, while structuralism views cultural artefacts (e.g. narrative texts) as systems governed by universal structures. Classical Narratology seeks to uncover deep narrative structures, identifying recurring patterns, functions, and roles such as character archetypes and plot sequences. 
    
    This concern with formal differentiation dates back to \citet{aristotle_poetics_penguin}’s Poetics, which distinguished the story world from \emph{mythos}, the aesthetic arrangement of events \citep{wang-etal-2021-learning-similarity}. Russian Formalism later introduced the contrast between \emph{fabula} (chronological events) and \emph{sujet} (their presentation) \citep{shklovsky1917art,tomaševskij1971theory}. Subsequent influential narratice theory works include \cite{propp1968morphology}’s Morphology of the Folktale, which identifies a fixed sequence of narrative functions. 
    
    French structuralism consolidated these ideas, giving rise to narratology as a systematic discipline. They coined the term \textit{narratologie} \cite{todorov1969grammaire}, the \bluetext{systematic study} of narrative, and reframed the duality of \emph{fabula} and \emph{sujet} as \emph{histoire} and \emph{récit} \citep{genette1980narrative}, terms rendered in English as \emph{story} and \emph{discourse} \citep{chatman1980story}. Across these shifts, the central aim remained the same: to classify narrative elements and the underlying structure of all narratives.

    During the 1980s and 1990s, narratology broadened its scope beyond literary traditions by integrating concepts from other disciplines, reflecting a wider shift in the humanities from structuralist to post-structuralist approaches \citep{Meister2014Narratology}. This transition marked the end of narratology’s “classical” phase, rooted in French structuralist and Russian formalist traditions that emphasised systematic models of narrative structure and discourse, and ushered in a ``postclassical” era. Postclassical narratology expands the field through interdisciplinary engagement with \redtext{the fields of} cognitive science, cultural studies, and media theory. \redtext{The expansion} incorporates \redtext{a new interest in }context, reception, and ideology, while \redtext{also retraining the existing core insights on textual form and structure} \citep{herman1997scripts, prince2008classical}. These developments gave rise to contextualist, cognitive, and other approaches\footnote{Transmedial/transgeneric narratology investigates how narrative structures and strategies function across different genres, including fiction, poetry, drama, and non-fictional discourse. It emphasises the adaptability of narrative beyond genre-specific norms, and considers how narrative techniques are reshaped in genres that foreground visual or multimodal conventions, such as graphic novels or visual poetry. As our survey focuses on textual narratives, we provide little to no coverage of LLM works involving transmedial/transgenric narratology.} that shaped \textbf{Postclassical Narratology} as a theoretically diverse and interdisciplinary practice:
    
    \begin{description}
        \item[\textbf{Contextualist Narratology:}] Contextual narratology examines narrative representation in relation to specific cultural, historical, ideological, and thematic contexts. It moves beyond structural analysis to consider how narratives reflect and engage with broader environments including gender, politics, and media.
        
        \item[\textbf{Cognitive Narratology:}] Cognitive narratology explores how narratives are mentally processed, understood, and represented by human agents. Drawing on cognitive science, it investigates how phenomena such as perspective-taking, empathy, narrative schemas, and the simulation of experience interact with the narrative text.
    \end{description}

\subsection{Narrative Levels}
\label{sec:narrative_levels}

    In this survey, we categorise LLM narrative system according to the narrative level they aim to represent/model through various understanding and generation tasks (Section \ref{sec:tasks_and_levels}). We build upon a tripartite model of narrative communication (story, discourse and narration) from the works of \cite{genette1980narrative, chatman1980story, gerald1973introduction}. Given that our target audience consists of NLP researchers, and that terms like \emph{story} and \emph{narrative} are often used interchangeably in computational contexts, we exchange the term \emph{story} for the other popular term \emph{fabula} to avoid confusion and adhere to the conventions of the field, similarly to \cite{gervas2024storytelling}. \bluetext{Furthermore, we adapt \cite{herman2009basic}'s postclasscial notion \emph{situatedness} as an addition to the tripartite model to address the considerations of contextual/cognitive narrative phenomena outside the text.}\footnote{\bluetext{This approach to taxonomy was similarly taken by \cite{hamilton-etal-2026-narrabench}.}} We use the terminology as follows:

    \begin{description}[itemsep=0.5em, labelwidth=!, leftmargin=0pt]
    
        \item[\textbf{Fabula:}] 
            The term \emph{fabula} refers to the chronological sequence of events that make up the story being told, abstracted from the mode of presentation (also called: histoire, story) It is the raw material of narrative, encompassing the actions, events, and situations in their temporal order. For example, \citet{hamilton-etal-2025-city} use LLMs to extract social networks from multilingual fiction and non-fiction narratives. We classify this as a \emph{fabula}-level representation.
    
        \item[\textbf{Discourse:}] 
            \emph{discourse} designates the narrative discourse: the actual telling or presentation of the story (also called: mythos, sujet, récit). It includes the structure, style, and medium through which \emph{fabula} is conveyed. Genette defines it as the text itself which shapes the raw events into a communicable form, involving choices of voice, focalisation, and temporal arrangement. For instance, \citet{xie-etal-2023-next} investigate the effect of varying prompt style, register, and length on GPT-3’s output, while \citet{troiano_clause-atlas_2024} use GPT-3.5-turbo to annotate clauses as events, subjective experiences, and contextual information. We classify both as \emph{discourse}-level representation.
        
        \item[\textbf{Narration:}] 
            The act or process of narrating, or the generating instance of narrative discourse \citep{genette1980narrative}. This involves the time of narration and the narrator's position relative to the diegesis (the story world): extradiegetic (e.g. a narrator who claims to tell a fictional story) or intradiegetic heterodiegetic (a narrator who is not a part of the story world) or homodiegetic (a narrator who is also a character in the story world). Narration also includes the presence of a \emph{narratee}: the entity to whom the narrative is addressed, whether explicitly or implicitly. The narrating situation thus comprises narrator, narratee, time, and person. For example, \citet{brei-etal-2025-classifying} use LLMs to assess narrator reliability, which we classify as a \emph{narration}-level representation.

        \item[\textbf{\bluetext{}{Situatedness:}}] 
            \bluetext{}{By synthesising insights from sociolinguistics, social psychology and narratology, \cite{herman2009basic} defines a `first' level of narrative which is that it must be situated in and interpreted in light of a specific context or occasion of telling. This is the focus of cognitive and contextualist narratology. For example, \citet{shen_heart-felt_2024} investigates the correlation between textual features and empathetic reactions from human readers, which is a phenomena in the situation outside the narrative text, which we classify as a \emph{situatedness}-level representation.}
    \end{description}

    Although these levels frequently overlap--and most surveyed NLP systems span more than one--defining them explicitly grounds computational methods in the theoretical foundations of narrative research, enabling systematic comparison across LLM-based approaches. Individual narrative theories and their applications are discussed in detail in Section \ref{sec_narrative_theories}.

\section{Related Work}
\label{sec_related_work}

We begin Section \ref{sec_related_work} by introducing early applications of theory-informed computational narrative modelling, before summarising related surveys published in the contemporary LLM era--which we position as beginning with the release of GPT-3 \citep{brown2020language}--and highlighting a notable gap in how emerging theory-driven LLM methods are addressed. \redtext{We then examine the categorisation systems used in these surveys, contrasting prior taxonomies with those structured around narrative theories \citep{piper2021narrative,wang2023open,gervas2024storytelling,hamilton-etal-2026-narrabench}, a trend our survey also follows.}


\subsection{\redtext{History}}

The earliest documented Automatic Story Generation (ASG) system was SAGA II \citep{ross1960saga}, which generated short Western playlets by using random probabilistic selection over a small set of pre-authored narrative branches \citep{sample2013account}. Subsequently, \citet{klein1973automatic}'s automatic novel-writing system generated murder mystery stories by maintaining a probabilistic semantic network representing characters, objects, and relations within a fictional world. A later approach was \citet{Meehan_talespin_1977}'s TALE-SPIN, a goal-directed problem solver in which simulated characters pursued plans to satisfy goals such as hunger or friendship; \redtext{it drew on \citet{schank1977scripts}'s broader framework of scripts, plans and goals, which would later influence NLP knowledge structures such as narrative event chains and schemas \citep{chambers-jurafsky-2008-unsupervised,chambers-jurafsky-2009-unsupervised}.}

While the aforementioned historical origins of ASG emerged independently of narrative theories, new evidence reveals that narrative theory-driven ASG methods were, in fact, in development from the start. \citet{ryan_grimes} details a previously unknown fairytale generation system developed by linguist Joseph E. Grimes around in either 1960 or 1961, at around the same time of the earliest documented SAGA II.

Grimes's system generated fairy tales by taking a grammar-based application of \citet{propp1968morphology}'s \emph{Morphology of the Folktale} by randomly selecting a subset of Propp's 31 narrative functions (such as `the hero leaves home' or `the villain is defeated') and arranging them in their proper structural sequence. It then generated simple English or Spanish prose by probabilistically selecting actions to fill pre-written templates. Grimes’s previously forgotten program stands as the earliest known precursor to the theory-driven ASG methods that would become prominent in the LLM era.

\subsection{Existing Surveys in the LLM Era}

\begin{table}[!htbp]
    \centering
    \footnotesize
    \rowcolors{2}{gray!10}{white}
    \begin{tabularx}{\textwidth}{@{}p{0.28\textwidth} | X | c | c | c | c@{}}
        \toprule
        Survey Paper & Journal/Venue & \rotatebox{90}{LLM methods} & \rotatebox{90}{Narrative Theories} & \rotatebox{90}{Generation} & \rotatebox{90}{Understanding} \\
        \midrule
        
        \cite{alabdulkarim-etal-2021-automatic} & Workshop of Narrative Understanding & \lmark & \xmark & \cmark & \xmark \\
        \cite{alhussain2021automatic} & ACM Computing Surveys & \xmark & \lmark & \cmark & \xmark \\
        \cite{piper2021narrative} & EMNLP & \xmark & \cmark & \xmark & \cmark \\
        
        \cite{berhe_survey_2022} & TAL Journal & \xmark & \lmark & \xmark & \cmark \\
        \cite{ranade2022computational} & IEEE Access & \xmark & \xmark & \cmark & \cmark \\
        
        \cite{keith2023survey} & ACM Computing Surveys & \xmark & \lmark & \xmark & \cmark \\
        \cite{santana2023survey} & AI Review & \xmark & \lmark & \xmark & \cmark \\
        \cite{wang2023open} & Neurocomputing & \lmark & \cmark & \cmark & \xmark \\
        
        \cite{fang2023systematic} & Education and Information Technologies & \xmark & \xmark & \cmark & \xmark \\
        \cite{hatzel2023machine} & Information Technology Journal & \lmark & \xmark & \xmark & \cmark \\
        \cite{piper-2023-computational} & The Big Picture Workshop & \xmark & \cmark & \xmark & \cmark \\
        \cite{zhu-etal-2023-nlp} & EMNLP & \cmark & \xmark & \xmark & \cmark \\

        \cite{gervas2024storytelling} & Interdisciplinary Science Reviews & \xmark & \cmark & \cmark & \cmark \\
        
        \cite{calvo2025integrating} & \bluetext{Mathematics} & \cmark & \xmark & \cmark & \xmark \\
        \cite{teleki-etal-2025-survey} & \bluetext{EMNLP} & \cmark & \xmark & \cmark & \xmark \\
        \cite{vissersolissa2025event} & \bluetext{Journal of Computational Literary Studies} & \lmark & \cmark & \xmark & \cmark \\

        \cite{hamilton-etal-2026-narrabench} & \bluetext{EACL} & \cmark & \cmark & \xmark & \cmark \\
        \cite{ma-etal-2026-text} & \bluetext{EACL} & \cmark & \xmark & \cmark & \xmark \\
        \cite{Mozaffari_literary_llm_survey} & \bluetext{Human Behaviour and Emerging Technologies} & \cmark & \xmark & \cmark & \cmark \\
        
        \midrule
        \textbf{Our Survey} & & \cmark & \cmark & \cmark & \cmark \\
        \botrule
    \end{tabularx}
    \caption{\redtext{Comparison of scopes/coverage from recent (post GPT-3) surveys of computational methods in narrative studies: \cmark\ Yes, \lmark\ Limited, \xmark\ No.}}
    \label{table:recent_surveys}
\end{table}




The growing preference towards LLM methods (Section \ref{sec_scope_trends}) is reflected in recent surveys (Table \ref{table:recent_surveys}) on automatic story generation and understanding. While older surveys such as \citet{ranade2022computational} investigated symbolic, statistical and non-LLM machine learning methods, \bluetext{newer surveys have mostly moved to prioritise LLM-based methods \citep{teleki-etal-2025-survey,Mozaffari_literary_llm_survey} or LLM benchmarks \citep{hamilton-etal-2026-narrabench}.} An exception to this trend is \citet{gervas2024storytelling}, who deliberately excludes machine learning methods, asserting that ``these efforts have had absolutely no input from the existing body of knowledge on narratology or any other discipline.'' \citet{piper2021narrative} hold a similar view: ``(prior NLP) is by and large divorced from the large body of theoretical work on narrative within the humanities, social and cognitive sciences." We examine these assertions by comparing existing surveys, providing an overview of their different categorisation methods while also motivating for our chosen approach in the current context\footnote{Our survey shows, through systematic examination of 68 papers published between 2021 and 2026, that the gap both \citet{piper2021narrative} and \citet{gervas2024storytelling} identify has narrowed considerably in the LLM era."}.

The disconnection between theories and NLP machine learning methods can be observed in many existing surveys, where engagement with narrative theories is missing or limited. Several surveys in Table \ref{table:recent_surveys} make little or no reference to narrative theories at all \citep{alabdulkarim-etal-2021-automatic, fang2023systematic, ranade2022computational} while others draw on narrative theories only to the extent of motivating or contextualising their scope \citep{alhussain2021automatic,berhe_survey_2022,santana2023survey}.

While our survey hopes to specifically connect theories to methods, we here point to existing surveys for readers with diverse intentions: for generation methods, \cite{alhussain2021automatic} and \cite{fang2023systematic} make the distinction between planning-based (which are more representative of symbolic methods) and statistical machine learning models in story generation, \bluetext{while \citet{teleki-etal-2025-survey} and \citet{ma-etal-2026-text}'s LLM-focused surveys categorise methods based off of whether humans are involved in the loop and single vs multi-agent systems.} In comparison, story understanding surveys are categorised by task, where \citet{santana2023survey} categorise the type of knowledge extracted--event, participants, time and space--while \citet{zhu-etal-2023-nlp} additionally distinguishes between abstractive vs extractive approaches for summarisation, and generation vs retrieval based approaches for question answering.

\redtext{We believe that a categorisation system which primarily depends on drawing distinctions between planning and machine learning, or between generation and understanding, is no longer sufficient as} LLM-enabled systems often combine these functionalities, outputs, and techniques. For example, an LLM-driven story generation system may simultaneously generate natural language text, maintain narrative representations using data structures, and infer ongoing character motivations. This system cannot be placed neatly into either taxonomy: it performs text generation and narrative understanding and/or interpretation (of the text being generated) while also using planning and machine learning. 

\redtext{In contrast, \citet{piper2021narrative} propose narrative theories as an organising framework for computational narrative research and argues that ``linking computational work in NLP to theory opens up a range of new empirical questions that would both help advance our understanding of narrative and open up new practical applications".} We found three surveys which have put this idea into practice by using narrative levels (Section \ref{sec:narrative_levels}) for categorisation. \citet{wang2023open} categorises open-world story generation with structured knowledge enhancement methods by their generation targets--story, plot and discourse (see Section \ref{sec_narrative_landscape} for terminology). Similarly, although \citet{gervas2024storytelling} surveys only symbolic approaches, he categorises them as fabula generation, discourse composition, narrative interpretation, and text
generation. \bluetext{\citet{hamilton-etal-2026-narrabench} survey LLM-benchmarks for narrative understanding and categorises tasks as story, discourse, narration and situatedness.}

\subsubsection{Filling the Gap}

In summary, the key gaps that motivated us to survey narrative theory-driven LLM methods for story understanding and generation are:

\begin{enumerate}
    \item \textbf{Lack of Integrated Frameworks:} Most existing surveys do not provide categorisations that align narrative theories with LLM methods.
    
    \item \textbf{Limited Understanding of Interactions:} The relationships between narrative theory, tasks, LLM methods, and datasets remain unexplored, including how these elements influence each other.
    
    \item \textbf{Insufficient Empirical and Strategic Insights:} There is limited analysis of patterns and trends in applying narrative theories to LLMs, clarity on task-theory alignment, and guidance for future research directions.
\end{enumerate}

Continuing the efforts of prior work, our present survey follows the spirit of using theory informed narrative levels as the primary organising principle for both the phenomena under investigation and the methods used to study them. This derives from the philosophy that ``linguistics is the eye and computation the body'' of NLP \citep{adityanlpbook}. Surveying the connection between theories and methods is motivated by the hypothesis that not only can LLM research incorporate insights from narratology and related disciplines, it also holds promise for contributing to interdisciplinary narrative studies through theory synthesis, validation, and new insights.

\section{Scope and Trends}\label{sec_scope_trends}

In Section \ref{sec_scope_trends}, we first explain our review methodology, which targets research which operationalises existing frameworks from narrative theories in their methodologies, either explicitly or implicitly. We then introduce the overall trends that we observed in the survey: the increased use of LLMs; more narrative theory-informed methods in story understanding than story generation; methods mostly focus on the fabula and discourse levels over situatedness, while the narration level is almost never addressed; theoretical contributions--including synthesis, taxonomy, and validation--are present in roughly a quarter to a third of surveyed papers; in-context approaches predominate most methods, while effective post-training and fine-tuning has been found for understanding tasks but is almost non-existent for generation tasks; and the surveyed papers mostly use datasets drawing from fictional or elicited narratives, compared to non-fictional or personal ones.

\subsection{\redtext{Review Methodology}}

\redtext{This survey is a scoping review, targeting representative coverage of NLP works in which narrative theories shape LLM methods.} We include research of methods for textual story generation and understanding where narrative theory is relevant to task formulation, label or prompt construction, model design, or output evaluation. Explicit citation of theory was not required: narratological concepts circulate beyond specialist scholarship and become embedded in everyday assumptions about storytelling \citep{mire-etal-2024-empirical}. A paper treating narrative as having ``a beginning, middle and end'' need not cite \citet{aristotle_poetics_penguin} to be drawing on a recognisable structural theory. A paper is treated as theory-driven when narrative concepts shaped its task definition, model design, prompting strategy, annotation scheme, representation, or evaluation criteria.

Categorisation requires interpretation. Narratological terms frequently appear without their lineage made explicit--\citet{beguvs2024experimental}, for example, invokes plot, conflict, point of view, and focalisation without tracing these to their narratological origins--while we found that explicit citations sometimes signal scope rather than guarantee substantial application. Attribution is further complicated by overlapping concepts across theories: \citet{yoo2024leveraging} attribute fabula and sujet to \citet{genette1980narrative}, though the terms originate in Russian formalism \citep{shklovsky1917art}. We therefore classify papers by weighing explicit citation against depth of application: papers that cite and meaningfully apply a theory are classified under it; papers that strongly operationalise a theory's core ideas without citing it are included as implicit applications. Classification reflects how a theory is used, not merely whether it appears in the reference list. This process is inherently subjective and our findings should be read as indicative of trends rather than definitive or exhaustive.

\redtext{While initial candidate papers were identified through keyword searches of the ACL Anthology using the terms `story' and `narrative', we actively expanded our corpus through backward and forward citation tracing from interdisciplinary surveys and workshops. This allowed us to capture 21 relevant literature extending beyond NLP and Computational Linguistics (Table \ref{tab:venue_discipline}).} We prioritised studies proposing, adapting, or evaluating LLM-based methods for textual narrative tasks, with priority given to decoder-only transformer models such as GPT and Llama. Selected encoder-only studies--such as BERT-based models--were included where their methods are directly and substantially shaped by theory-informed narrative representation, annotation, or evaluation, as such models remain widely used alongside LLM pipelines. Work relying primarily on non-LLM approaches--symbolic systems, statistical models, convolutional neural networks, or image and audio generation--was excluded.

\redtext{We include only work in which a computational method is itself developed, evaluated, or compared. CHR and DH research is discussed where it intersects with method development; purely qualitative studies, which typically present methods as proofs of concept without the systematic evaluation characteristic of NLP, fall outside the survey's scope. In total, 68 papers were collected; Table \ref{tab:venue_discipline} summarises their distribution by venue and discipline.}


    \begin{table}[h]
        \centering
        \begin{tabular}{lc}
        \toprule
        \textbf{Discipline} & \textbf{Papers} \\
        \midrule
        Computational Linguistics / NLP & 47 \\
        Digital Humanities               & 9  \\
        Artificial Intelligence          & 5  \\
        Humanities, Social \& Cognitive Sciences & 4  \\
        Human--Computer Interaction      & 3  \\
        \midrule
        \textbf{Total}                   & \textbf{68} \\
        \bottomrule
        \end{tabular}
        \caption{\redtext{Distribution of surveyed papers by discipline inferred from publication venue.}}
    \label{tab:venue_discipline}
    \end{table}
    
\subsection{Increasing Use of LLMs in Narrative Understanding Workshops}

    \begin{figure}[!htbp]
        \centering
        \resizebox{0.8\textwidth}{!}{%
            \begin{tikzpicture}
                \begin{axis}[
                    width=\textwidth,
                    height=6cm,
                    xlabel={Workshop Year},
                    ylabel={Number of Papers},
                    xtick={2021,2022,2023,2024,2025},
                    ytick={0,2,4,6,8,10},
                    xmin=2021, xmax=2025,
                    enlarge x limits=false, 
                    grid=major,
                    tick label style={font=\small,/pgf/number format/1000 sep=}, 
                    label style={font=\small},
                    legend style={font=\small, at={(0.5,1.05)}, anchor=south, legend columns=2}
                ]
                    \addplot[thick, color=blue, mark=*]
                        coordinates {(2021,2) (2022,3) (2023,0) (2024,4) (2025,10)};
                    \addlegendentry{\redtext{}{Uses LLM} (e.g. GPT)}
                    
                    \addplot[thick, color=red, mark=square*]
                        coordinates {(2021,5) (2022,4) (2023,4) (2024,1) (2025,0)};
                    \addlegendentry{\redtext{No LLM but uses }Encoder (e.g. BERT)}
                    
                    \addplot[thick, color=black, mark=triangle*]
                        coordinates {(2021,0) (2022,1) (2023,7) (2024,2) (2025,0)};
                    \addlegendentry{\redtext{No LLM} (e.g. LSTM, statistical, symbolic)}
                \end{axis}
            \end{tikzpicture}%
        }
        \caption{\redtext{Methods} used in the papers selected by the Workshop on Narrative Understanding. Proceedings: \url{https://aclanthology.org/venues/wnu/}}
        \label{fig:Workshop_NU}
    \end{figure}
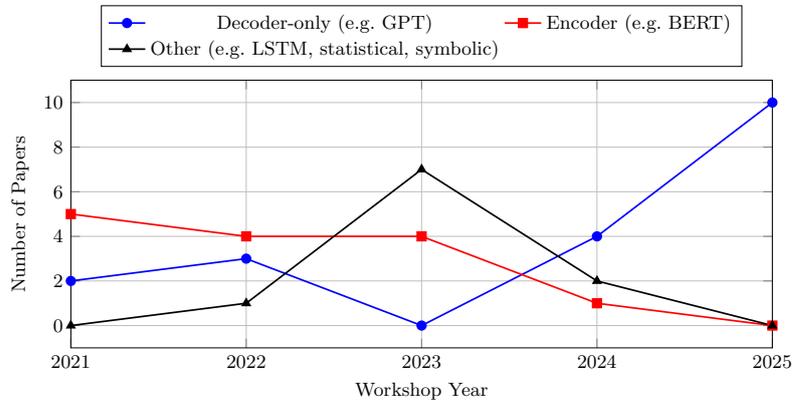

    LLMs outperform earlier methods on automatic story generation in terms of fluency, coherence, and task-specific objectives such as maintaining plot consistency, generating longer narratives, and improving controllability \citep{wang2023open}. At story understanding tasks, LLMs surpass earlier methods in identifying subtle thematic patterns and connecting narratives to broader cultural or historical contexts such as myths, religion, and history \citep{piper-wu-2025-evaluating}. They are also found to be more generalisable across narrative contexts compared to encoder-only models \citep{piper_narradetect_2025}.
    
    These performance advantages have driven a noticeable shift in methodological preferences within the narrative research community. We investigate this trend through the Workshop on Narrative Understanding (WNU), an annual interdisciplinary event bringing together researchers from AI, ML, NLP, and the humanities to advance automatic narrative understanding and generation (see Figure \ref{fig:Workshop_NU}). In 2025, 10 out of 10 accepted papers used a decoder-only model (e.g. GPT, Llama), compared to 4 out of 9 in 2024--suggesting that decoder-only LLMs are becoming the default choice in a space that previously favoured encoder-only models.

\subsection{Categorisation Methodology}

To position an LLM-based narrative method within our framework, we consider six variables:

\begin{enumerate}[leftmargin=1.4em, itemsep=2pt, topsep=2pt]
    \item \textbf{Theoretical Inspiration:} the theor(ies) which informed the task and/or implementation of the method, whether implicitly or explicitly applied (Section \ref{sec_narrative_theories}).
    \item \textbf{Data source:} fiction, non-fiction, personal, or elicited\footnote{\bluetext{Elicitation is a technique where the researcher provides a stimulus to a speaker to produce a specific response under experimental conditions, distinguishing it from the collection of primary texts (natural discourse). \citep{chelliah2010handbook,biber1993representativeness}.}}
        (Section \ref{sec:narrative_sources}).
    \item \textbf{Narrative level:} fabula, discourse, narration and situatedness. A method may target more than one (Section \ref{sec:tasks_and_levels}).
    \item \textbf{Understanding/Generation/Both:} whether the goal fo the system is narrative understanding, generation or both.
        (Section \ref{sec_modeling}).
    \item \textbf{LLM Method:} in-context, post-training, or compound systems.
        (Section~\ref{sec_LLM_techniques}).
    \item \textbf{Contributions:} resource contributions (Section X) or theoretical contributions (Section \ref{subsec:theoretical_contributions}).
\end{enumerate}
    
\subsection{Trends in Papers Surveyed}


\begingroup
    \footnotesize
    \rowcolors{2}{gray!10}{white}
    \setlength{\tabcolsep}{2pt}
    \renewcommand{\arraystretch}{1.1}
    
    \begin{xltabular}{\textwidth}{@{} >{\raggedright\arraybackslash}p{3cm} | *{4}{>{\centering\arraybackslash}X|} *{4}{>{\centering\arraybackslash}X|} *{2}{>{\centering\arraybackslash}X|} *{3}{>{\centering\arraybackslash}X|} *{4}{>{\centering\arraybackslash}X|} >{\centering\arraybackslash}X @{}}
    
        \caption{Narrative source, level, task, methods and contributions of surveyed research}
        \label{tab:focus_table} \\
        \toprule
        \textbf{Reference} & 
        \multicolumn{4}{c|}{\textbf{Source}} & 
        \multicolumn{4}{c|}{\textbf{Level}} & 
        \multicolumn{2}{c|}{\textbf{Task}} & 
        \multicolumn{3}{c|}{\textbf{LLM}} & 
        \multicolumn{5}{c}{\textbf{Outcome}} \\
        
        \cmidrule(lr){1-1} \cmidrule(lr){2-5} \cmidrule(lr){6-9} \cmidrule(lr){10-11} \cmidrule(lr){12-14} \cmidrule(lr){15-19}
        
        & 
        \rot{Fiction} & \rot{Non-fiction} & \rot{Personal} & \rot{Elicited} 
        & \rot{Fabula} & \rot{Discourse} & \rot{Narration} & \rot{Situatedness} 
        & \rot{Understanding} & \rot{Generation} 
        & \rot{In Context} & \rot{Training} & \rot{Compound} 
        & \rot{Dataset} & \rot{Benchmark} & \rot{Synthesis*} & \rot{Taxonomy*} & \rot{Validation*} \\
        \midrule
        \endfirsthead
        
        \caption[]{Narrative source, level, task, methods and contributions of surveyed research (Continued)} \\
        \toprule
        \textbf{Reference} & 
        \multicolumn{4}{c|}{\textbf{Source}} & 
        \multicolumn{4}{c|}{\textbf{Level}} & 
        \multicolumn{2}{c|}{\textbf{Task}} & 
        \multicolumn{3}{c|}{\textbf{LLM}} & 
        \multicolumn{5}{c}{\textbf{Outcome}} \\
        
        \cmidrule(lr){1-1} \cmidrule(lr){2-5} \cmidrule(lr){6-9} \cmidrule(lr){10-11} \cmidrule(lr){12-14} \cmidrule(lr){15-19}
        
        & 
        \rot{Fiction} & \rot{Non-fiction} & \rot{Personal} & \rot{Elicited} 
        & \rot{Fabula} & \rot{Discourse} & \rot{Narration} & \rot{Situatedness} 
        & \rot{Understanding} & \rot{Generation} 
        & \rot{In Context} & \rot{Training} & \rot{Compound} 
        & \rot{Dataset} & \rot{Benchmark} & \rot{Synthesis*} & \rot{Taxonomy*} & \rot{Validation*} \\
        \midrule
        \endhead
        
        \bottomrule
        \multicolumn{19}{r}{\textit{Continued on next page...}} \\
        \endfoot
        
        \bottomrule
        \endlastfoot
    











        \cite{wilmot2021memory}
        & \checkmark &  & & 
        & \checkmark& \checkmark& &
        & \checkmark &
        & & \checkmark & \checkmark
        & &\checkmark & & &\checkmark\\

        \cite{gangal2022nareor}
        & & & & \checkmark    
        & \checkmark & \checkmark & &
        &  & \checkmark
        & & \checkmark & 
        & \checkmark&\checkmark & & & \\

        \cite{hamilton2022covid}
        & & \checkmark & &     
        & \checkmark & \checkmark & \checkmark & \checkmark
        & & \checkmark
        & & \checkmark & 
        & & & & & \checkmark \\
        
        \cite{mori2022plug}
        & & & & \checkmark
        & \checkmark& \checkmark& &
        &  & \checkmark
        & & \checkmark&\checkmark
        &&&&&\\


        \cite{xie2022psychology}
        && &  & \checkmark
        & \checkmark& \checkmark& &
        &  & \checkmark
        & & \checkmark & \checkmark
        & \checkmark & & & & \\

        \cite{bizzoni-feldkamp-2023-comparing}
        & \checkmark & & &
        & & \checkmark &&
        & \checkmark &
        & & &\checkmark
        & \checkmark & \checkmark &  & & \\

        \cite{mirowski2023co}
        & & & & \checkmark
        & \checkmark& \checkmark& &
        & & \checkmark
        & \checkmark &  & \checkmark
        & & & & & \\

        \cite{shen-etal-2023-modeling}
        & & \checkmark& \checkmark& \checkmark
        & \checkmark& \checkmark& \checkmark&\checkmark
        & \checkmark &
        & \checkmark&\checkmark&
        & \checkmark &\checkmark & & & \\

        \cite{xie-etal-2023-next}
        &  \checkmark & \checkmark &  & \checkmark
        &  \checkmark & \checkmark & &
        & & \checkmark
        & \checkmark & \checkmark & \checkmark
        & \checkmark & \checkmark & & & \\

        \cite{zhao2023parrot}
        &  \checkmark & &  &
        &  \checkmark & \checkmark & &
        & \checkmark & 
        & & \checkmark & 
        & & & & & \\

        \cite{abdessamed2024identifying}
        & \checkmark & \checkmark & \checkmark & 
        &  \checkmark & \checkmark & &
        & \checkmark &
        & \checkmark &&
        & \checkmark & & & & \checkmark \\

        \cite{antoniak-etal-2024-people}
        &  \checkmark & \checkmark & \checkmark & 
        & \checkmark & \checkmark & \checkmark &
        & \checkmark & 
        & \checkmark & \checkmark &
        & \checkmark & \checkmark & & & \\

        \cite{bartalesi2024using}
        & \checkmark & \checkmark & & \checkmark
        & \checkmark & \checkmark & &
        & \checkmark &
        & \checkmark & &
        & & & & & \\

        \cite{beguvs2024experimental}
        & \checkmark & \checkmark & \checkmark & \checkmark
        & \checkmark & \checkmark & & \checkmark
        & & \checkmark
        & \checkmark & &
        & \checkmark & & & & \checkmark \\
        
       \cite{brei2024returning}
        & & & & \checkmark   
        & \checkmark & \checkmark & &
        & \checkmark & \checkmark
        & \checkmark&\checkmark &\checkmark 
        & & & &&\checkmark \\

        \cite{chakrabarty2024art}
        & \checkmark & & & \checkmark
        & \checkmark & \checkmark & \checkmark &
        & \checkmark & \checkmark
        & \checkmark & &
        & \checkmark & \checkmark & & \checkmark & \\

        \cite{chun_aistorysimilarity_2024}
        & \checkmark & & &
        & \checkmark & \checkmark & &
        & \checkmark &
        & \checkmark & &
        & \checkmark & \checkmark & & \checkmark & \\

        \cite{das-etal-2024-media}
        & & \checkmark & &    
        & \checkmark & \checkmark &  & \checkmark 
        & \checkmark & 
        & \checkmark & \checkmark & \checkmark 
        & \checkmark & & & & \\
        
        \cite{ermolaeva-etal-2024-tame}
        & \checkmark & & & \checkmark
        & \checkmark & \checkmark &  &  
        &  &  \checkmark
        & \checkmark & &  
        & & & & & \\

        \cite{gervas2024tagging}
        & \checkmark& & &    
        & \checkmark& \checkmark&&
        & \checkmark &
        & \checkmark&\checkmark & 
        &&&&&\\

        \cite{halperin2024artificial}
        & & & & \checkmark
        & \checkmark & \checkmark && 
        & & \checkmark
        & \checkmark & & \checkmark
        & \checkmark & & & & \\

        \cite{harel-canada-etal-2024-measuring}
        & \checkmark &&& \checkmark
        & \checkmark & \checkmark &  & \checkmark
        & \checkmark & \checkmark
        & \checkmark & &
        & \checkmark & & \checkmark & \checkmark & \\

        \cite{hatzel-biemann-2024-story}
        & \checkmark & & & \checkmark
        & \checkmark & \checkmark & &
        & \checkmark &
        & \checkmark & \checkmark &
        & \checkmark & & & & \\

        \cite{hobson-etal-2024-story}
        & \checkmark& \checkmark & \checkmark & 
        & & \checkmark & \checkmark & \checkmark
        & \checkmark &
        & \checkmark & &
        & \checkmark & & \checkmark & \checkmark & \\

        \cite{heddaya-etal-2024-causal}
        & & \checkmark & &    
        & \checkmark& \checkmark& & \checkmark
        & \checkmark &
        & \checkmark&\checkmark & 
        & \checkmark&&\checkmark& & \\

        \cite{lima2024pattern}
        & \checkmark& & &    
        & \checkmark& \checkmark&&
        & \checkmark & \checkmark
        & \checkmark& &\checkmark
        & & && & \\
        
        \cite{marino2024integrating}
        & & \checkmark & &
        & & \checkmark & & \checkmark
        & \checkmark &
        & \checkmark & \checkmark & \checkmark
        & & & & & \\

        \cite{mire-etal-2024-empirical}
        & & \checkmark & \checkmark & 
        &  \checkmark& \checkmark & \checkmark & \checkmark
        & \checkmark &
        & \checkmark & &
        & \checkmark & & & \checkmark & \checkmark \\
    
        \cite{piper-bagga-2024-using}
        & \checkmark & \checkmark & \checkmark &
        & & \checkmark & &
        & \checkmark &
        & \checkmark & \checkmark &
        & \checkmark & & \checkmark & \checkmark & \checkmark \\

        \cite{piper-etal-2024-social}
        & \checkmark & \checkmark & &
        & \checkmark & & & \checkmark
        & \checkmark &
        & \checkmark & \checkmark & \checkmark
        & \checkmark & & \checkmark & & \checkmark \\

        \cite{shen_heart-felt_2024}
        & & \checkmark & \checkmark & \checkmark
        & \checkmark & \checkmark & \checkmark & \checkmark
        & \checkmark &
        & \checkmark & &
        & \checkmark & & \checkmark & \checkmark & \checkmark \\

        \cite{subbiah2024reading}
        & & \checkmark & \checkmark & \checkmark
        & \checkmark & \checkmark & \checkmark & \checkmark
        & & \checkmark
        & \checkmark & & \checkmark
        & \checkmark &\checkmark && \checkmark & \\

        \cite{song2024conflict}
        & &  &  & \checkmark
        & \checkmark& \checkmark&  &
        & \checkmark & \checkmark
        & \checkmark & \checkmark & \checkmark
        & & && & \\

        \cite{sui_confabulation_2024}
        & & & & \checkmark
        & &\checkmark & &
        & \checkmark &
        & & \checkmark &
        & & & \checkmark & & \checkmark \\

         \cite{sun-etal-2024-event}
        & & & & \checkmark
        & \checkmark & & &
        & \checkmark &
        & \checkmark & & \checkmark
        & & & & & \\

        \cite{tian-etal-2024-large-language}
        & \checkmark & & & \checkmark
        & \checkmark & \checkmark & &
        & \checkmark & \checkmark
        & \checkmark & &
        & \checkmark & \checkmark & \checkmark & \checkmark & \\
        
        \cite{troiano_clause-atlas_2024}
        & \checkmark & & &
        & \checkmark & \checkmark & &
        & \checkmark &
        & \checkmark & &
        & \checkmark & & & & \\

        \cite{vykopal2024disinformation}
        & & &\checkmark & \checkmark
        & \checkmark& \checkmark&  &\checkmark
        & \checkmark & \checkmark
        & \checkmark &  &
        & \checkmark & & & & \\

        \cite{yang-anderson-2024-evaluating}
        & \checkmark & & &
        & \checkmark & \checkmark &  &  
        & \checkmark &  
        & \checkmark & &  
        & \checkmark & \checkmark & & \checkmark & \\

        \cite{yoo2024leveraging}
        & & & & \checkmark    
        & \checkmark& \checkmark&&
        & \checkmark & \checkmark 
        & \checkmark&&\checkmark 
        & & & & & \\

        \cite{zhou-etal-2024-large}
        & & \checkmark & &
        & \checkmark & \checkmark & \checkmark & \checkmark
        & \checkmark &
        & \checkmark & &
        & & & \checkmark & \checkmark & \\

        \cite{ahuja2025finding}
        & \checkmark & & & 
        & \checkmark & \checkmark & &
        & \checkmark & \checkmark
        & \checkmark & & \checkmark
        & \checkmark & \checkmark & & & \\

        \cite{brei-etal-2025-classifying}
        & \checkmark & \checkmark & \checkmark & 
        & \checkmark & \checkmark & \checkmark & \checkmark
        & \checkmark &
        & \checkmark & \checkmark &
        & \checkmark & & & & \\

        \cite{ghaffari-hokamp-2025-narrative}
        &&&& \checkmark  
        & \checkmark & \checkmark & &
        &  & \checkmark
        & \checkmark & & \checkmark
        &&&&&\\

        \cite{nikolaidis-etal-2025-polynarrative}
        & & \checkmark & & 
        & & \checkmark & & \checkmark
        & \checkmark &
        & \checkmark &  &
        & \checkmark & \checkmark & &\checkmark & \\

        \cite{sancheti-rudinger-2025-tracking}
        & \checkmark & & &    
        & \checkmark&\checkmark & &
        & \checkmark & 
        & \checkmark& &\checkmark 
        & & & &\checkmark & \\

        \cite{yu-etal-2025-multi}
        & \checkmark & & & \checkmark
        & \checkmark & \checkmark & &
        & & \checkmark
        & & & \checkmark
        & & & & & \\

        \cite{bissell-etal-2025-theoretical}
        & \checkmark& & &    \checkmark
        & \checkmark& \checkmark&&\checkmark
        & \checkmark & \checkmark
        & \checkmark&& 
        & \checkmark&  & \checkmark&\checkmark&\checkmark \\

        \cite{hamilton-etal-2025-city}
        & \checkmark & \checkmark &  & \checkmark
        & \checkmark & & &
        & \checkmark &
        & \checkmark & &
        & \checkmark & & & & \checkmark \\

        \cite{hobson2025evaluating}
        & & \checkmark &  & 
        & \checkmark & \checkmark & \checkmark & \checkmark
        & \checkmark &
        & \checkmark & &
        & \checkmark & & & & \checkmark \\

        \cite{liu2025raemollm}
        & \checkmark & & & \checkmark
        & & \checkmark & & \checkmark
        & \checkmark & 
        & \checkmark & & \checkmark
        & &&&& \\

        \cite{mitran2025probing}
        & \checkmark & & & 
        & \checkmark & \checkmark & \checkmark & \checkmark
        & \checkmark & 
        & \checkmark & & 
        & &&&& \checkmark \\

        \cite{piper_narradetect_2025}
        & \checkmark& \checkmark & \checkmark & 
        & \checkmark& \checkmark& \checkmark &
        & \checkmark & 
        & \checkmark & \checkmark &
        & \checkmark&\checkmark &\checkmark &\checkmark &\checkmark \\

        \cite{piper2025cr4}
        & \checkmark & \checkmark & & 
        & \checkmark & \checkmark &  & \checkmark
        & \checkmark & 
        & \checkmark & &
        & \checkmark & \checkmark & & & \\

        \cite{piper-wu-2025-evaluating}
        & \checkmark & \checkmark & &
        & \checkmark & & &
        & \checkmark &
        & \checkmark & & \checkmark 
        & \checkmark & & & & \\


        \cite{ran-etal-2025-bookworld}
        & \checkmark & & &
        & \checkmark & \checkmark & &
        & \checkmark & \checkmark
        & \checkmark & & \checkmark
        & & & & & \\

        \cite{shen2025adapting}
        & \checkmark && &    
        & \checkmark& \checkmark&  &
        & \checkmark & 
        & \checkmark&\checkmark& 
        & \checkmark&&&\checkmark &\\

        \cite{udrea2025speaks}
        & \checkmark&&&    
        &\checkmark&\checkmark&&
        & \checkmark & 
        &&&\checkmark 
        & \checkmark& & \checkmark&&\checkmark \\

        \cite{wang2025generating}
        &&& &    
        &\checkmark&\checkmark&&
        & \checkmark &\checkmark
        &&&\checkmark 
        &&&&&\\

        \cite{yang-jin-2025-matters}
        & \checkmark & & & 
        & \checkmark & \checkmark & & \checkmark
        & \checkmark &
        & \checkmark & \checkmark &
        & \checkmark & \checkmark & & \checkmark & \checkmark \\

        \cite{xia2025storywriter}
        & &&&
        & \checkmark& \checkmark&&
        & & \checkmark
        &&\checkmark&\checkmark 
        &\checkmark&&&&\\

        \cite{brei2026casper}
        & \checkmark&&&\checkmark    
        & \checkmark&\checkmark& &
        &\checkmark& 
        &\checkmark&&
        &\checkmark&&\checkmark&\checkmark&\checkmark \\

        \cite{sterner-etal-2026-contrastive} 
        & & & & \checkmark 
        & \checkmark & \checkmark &  &  
        & \checkmark &  
        & \checkmark & \checkmark &  
        & \checkmark & \checkmark & \checkmark &  & \checkmark \\

        \cite{liu2026retell}
        &&&&\checkmark
        &\checkmark&\checkmark&&
        & \checkmark&\checkmark 
        & \checkmark&\checkmark & 
        & \checkmark&\checkmark&\checkmark&\checkmark&\checkmark \\

        \cite{ma2026multi}
        & & &\checkmark &\checkmark    
        & \checkmark& \checkmark&\checkmark&\checkmark
        & \checkmark &  
        &&&\checkmark
        &&&\checkmark&\checkmark&\checkmark\\

        \cite{maggini2026partisanlens}
        & & \checkmark & &   
        &  & \checkmark &  &  \checkmark
        & \checkmark &  
        & \checkmark & \checkmark &  
        & \checkmark & \checkmark &  & \checkmark & \checkmark \\

        \cite{ghosal2026ai}
        & & & & \checkmark   
        & \checkmark & \checkmark &  &  \checkmark
        & & \checkmark  
        & \checkmark &  &  
        & & & & & \checkmark\\

        \cite{schneider2026llm}
        & &&\checkmark &    
        & \checkmark&\checkmark&&\checkmark
        & \checkmark & 
        & \checkmark& &\checkmark
        & \checkmark& & \checkmark&\checkmark &\checkmark\\


    \end{xltabular}
    
\endgroup

\begin{table}[h]
    \centering
    \small
    \begin{tabular}{lrrrr}
    \toprule
    \textbf{Category} & \textbf{Total} & \textbf{Understanding} & \textbf{Generation} & \textbf{Both} \\
    \midrule
    \textit{Task totals} & 68 & 41 & 14 & 13 \\
    \midrule
    \textit{Source}     &    &    &    &    \\
    \quad Fiction       & 37 & 26 &  4 &  7 \\
    \quad Non-fiction   & 25 & 21 &  4 &  0 \\
    \quad Personal      & 14 & 11 &  2 &  1 \\
    \quad Elicited      & 32 & 11 & 12 &  9 \\
    \midrule
    \textit{Level}      &    &    &    &    \\
    \quad Fabula        & 60 & 33 & 14 & 13 \\
    \quad Discourse     & 68 & 41 & 14 & 13 \\
    \quad Narration     & 14 & 11 &  2 &  1 \\
    \quad Situatedness  & 26 & 19 &  4 &  3 \\
    \midrule
    \textit{LLM Method} &    &    &    &    \\
    \quad In-Context    & 55 & 35 &  8 & 12 \\
    \quad Training      & 27 & 18 &  6 &  3 \\
    \quad Compound      & 28 & 12 &  9 &  7 \\
    \midrule
    \textit{Outcome}    &    &    &    &    \\
    \quad Dataset       & 43 & 29 &  7 &  7 \\
    \quad Benchmark     & 19 & 12 &  3 &  4 \\
    \quad Synthesis*    & 17 & 13 &  0 &  4 \\
    \quad Taxonomy*     & 22 & 16 &  1 &  5 \\
    \quad Validation*   & 24 & 18 &  3 &  3 \\
    \bottomrule
    \end{tabular}
    \caption{\redtext{Occurrence counts across surveyed NLP research ($n = 68$). Papers may
    appear in multiple sub-categories. Asterisked outcomes
    (*) indicate theoretical contributions.}}
    \label{tab:summary_counts}
\end{table}

\begin{enumerate}[label=\textbf{F\arabic*.}, leftmargin=2.2em, itemsep=4pt]

  \item \textbf{\redtext{Methodological differences in generation and understanding.}}
        We found 18 implementations of training-based methods for understanding tasks, compared to only 6 for generation tasks, of which 4 incorporate training as a subcomponent of a compound system (e.g., RAG). Only three papers apply training exclusively for generation as a stand-alone method \citep{gangal2022nareor, hamilton2022covid, liu2026retell}. Established, dedicated theory-informed fine-tuning methods for open-ended narrative generation are rare (Section~\ref{finetune}). The dominance of elicited data in generation tasks ($n = 21$) reflects a continued reliance on purpose-built, constrained datasets rather than naturally occurring text.

  \item \textbf{\redtext{The narration level is largely unaddressed, and situatedness is
        domain-bound.}}
        Narration is the least-explored narrative level in the surveyed papers ($n = 14$), with only three generation papers targeting it. Situatedness is addressed more ($n = 26$) but its involvement with generation tasks is limited ($n = 7$), and it appears predominantly in non-fictional tasks. 17 of 26 situatedness papers draw on non-fictional sourcess, covering misinformation, news framing, and political discourse. It is almost entirely
        absent from works that generate stories in the style of literary fiction (Section~\ref{sec:tasks_and_levels}).

  \item \textbf{\redtext{Literary narrative theories are being applied beyond their
        original contexts.}}
        The theoretical focus of surveyed work draws predominantly from literary narratology, yet the datasets it is applied to span a wide range of non-literary texts: non-fictional sources appear in 25 papers, elicited narratives in 32, and personal narratives in 14. The application of theories developed for literary narrative to journalism \citep{huang-usbeck-2024-narration,zhou-etal-2024-large}, social media \citep{antoniak-etal-2024-people} and political communication \citep{liu2025raemollm} represents an emerging research direction where these frameworks travel across domains.

  \item \textbf{\redtext{A meaningful proportion of papers contribute back to narrative
        theory.}}
        While dataset construction remains the most common research outcome ($n = 43$), explicit theoretical contributions are also prevalent: synthesis, taxonomy, and validation appear in roughly a quarter to a third of papers we surveyed ($n = 17$, $22$, and $24$ respectively). These contributions are concentrated in understanding tasks, suggesting that understanding tasks are currently the primary vehicle through which NLP research engages in the bidirectional relationship with narrative theory (Section~\ref{subsec:theoretical_contributions}).

\end{enumerate}

\section{Applications of Narrative Theories}
\label{sec_narrative_theories}

We use \cite{herman2009basic}'s fabula-discourse-narration-situatedness distinction to categorise narrative theories applied by LLM enabled narrative systems (Table \ref{tab:NT_table}). Unlike early symbolic approaches that focused exclusively on formalist theories to create grammar based systems \citep{alhussain2021automatic,wang2023open}, modern LLM narrative research demonstrates a greater variety of theoretical basis (\redtext{T}able \ref{tab:NT_table}). At the same time, we explore the trends in the relationship between narrative theories and their relevant tasks in NLP research. \redtext{We find that classical narratological theories dominate across both understanding and generation tasks, but generation research draws almost exclusively from formalist and structuralist frameworks, whereas understanding research engages a broader theoretical base including cognitive, contextual, and social science approaches.}

\newcolumntype{Y}{>{\raggedright\arraybackslash}X} 
\newcolumntype{S}[1]{>{\raggedright\arraybackslash}p{#1}} 
        
\begin{table}[!htbp]
\footnotesize
\centering
\setlength{\tabcolsep}{3pt}
\begin{tabular}{p{4.5cm} | c c c c | p{2cm} | p{4.2cm}}
\toprule
\textbf{Narrative Theory}
  & \rotatebox{90}{\textbf{Fabula}}
  & \rotatebox{90}{\textbf{Discourse}}
  & \rotatebox{90}{\textbf{Narration}}
  & \rotatebox{90}{\textbf{Situatedness}}
  & \textbf{Origin}
  & \textbf{Application} \\
\midrule
Poetics\newline\cite{aristotle_poetics_penguin} & \checkmark & \checkmark & & & Ancient Greece &
(G) \citet{mirowski2023co}
\\
\hline
Freytag’s Pyramid\newline\cite{freytag1894technik} & \checkmark & \checkmark & & & Drama &
(G) \citet{song2024conflict}
\\
\hline
Fabula/Sujet\newline\citep{shklovsky1917art} & \checkmark & \checkmark & & & Russian\newline Formalism &
(G) \citet{yu-etal-2025-multi}\newline
(G) \cite{yoo2024leveraging} \newline
\\
\hline
Morphology of the Folktale\newline\citep{propp1928morfologiya}& \checkmark & \checkmark & & & Russian\newline Folktales &
(U) \citet{hobson-etal-2024-story,hobson2025evaluating}\newline
(U) \citet{gervas2024tagging} \newline
(G) \citet{ermolaeva-etal-2024-tame} \newline
(G) \cite{halperin2024artificial}
\\
\hline
Hero's Journey (Monomyth)\newline\citep{Campbell1949Hero} & \checkmark & \checkmark & & & Mythology &
(G) \citet{lima2024pattern}\newline
(G) \citet{wang2025generating}
\\
\hline
Narrative Equilibrium\newline\citep{todorov19712} & \checkmark & \checkmark & & & \textbf{Classical\newline Narratology} &
(U) \citet{heddaya-etal-2024-causal}\newline
(B) \citet{liu2026retell}\newline
(B) \citet{sterner-etal-2026-contrastive}
\\
\hline
Cardinal Functions/Nuclei\newline\citep{barthes1975introduction} & \checkmark & \checkmark & & & \textbf{Classical\newline Narratology} &
(U) \citet{wilmot2021memory}\newline
(U) \citet{huang-usbeck-2024-narration}\newline
(U) \citet{sterner-etal-2026-contrastive}
\\
\hline
Shapes of Stories\newline\citep{vonnegut_shapes_of_stories} & \checkmark & \checkmark & & & Anthropology\newline (Originally A Rejected Masters Thesis) &
(G) \cite{mori2022plug}\newline
(G) \cite{xie2022psychology}\newline
(G) \citet{ermolaeva-etal-2024-tame}\newline
(B) \citet{tian-etal-2024-large-language} \newline
(U) \cite{bizzoni-feldkamp-2023-comparing}
\\
\hline
Distant Reading and \newline Network Theory\newline\citep{moretti2005graphs,moretti2011network} & \checkmark & \checkmark & & & Literary\newline Studies&
(U) \citet{yang-anderson-2024-evaluating}\newline
(U) \citet{sancheti-rudinger-2025-tracking}\newline
(U) \cite{piper-etal-2024-social}\newline
(U) \cite{udrea2025speaks}
\\
\hline
Rhetoric of Fiction\newline\citep{booth1961rhetoric} & \checkmark & \checkmark & \checkmark & & \textbf{Classical\newline Narratology} &
(U) \citet{hobson-etal-2024-story}\newline
(U) \citet{zhou-etal-2024-large}\newline
(U) \citet{brei-etal-2025-classifying}
\\
\hline
(Oral) Personal Experience\newline\cite{labov1967narrative} & \checkmark & \checkmark & \checkmark & & Sociolinguistics &
(U) \citet{shen-etal-2023-modeling}\newline
(U) \citet{troiano_clause-atlas_2024}\newline
(U) \citet{sterner-etal-2026-contrastive}\newline
(U) \citet{ma2026multi}\newline
(U) \citet{watahiki2026developing}
\\
\hline
Narrative Communication\newline\citep{chatman1980story} & \checkmark & \checkmark & \checkmark & & \textbf{Classical\newline Narratology} &
(U) \citet{hamilton-2024-detecting}\newline
(U) \citet{shen2025adapting}\newline
(U) \citet{brei-etal-2025-classifying}\newline
(U) \citet{chakrabarty2024art}
\\
\hline
Narrative Discourse\newline\citep{genette1980narrative} & \checkmark & \checkmark & \checkmark & & \textbf{Classical\newline Narratology} &
(G) \cite{gangal2022nareor}\newline
(U) \cite{zhao2023parrot}\newline
(U) \citet{piper-bagga-2024-using}\newline
(U) \citet{hatzel-biemann-2024-story}\newline
(G) \cite{ermolaeva-etal-2024-tame}\newline
(G) \cite{xia2025storywriter}
\\
\hline

Narrative Universals\newline\citep{sternberg1990telling} & \checkmark & \checkmark & \checkmark & \checkmark & \textbf{Postclassical\newline Narratology} &
(U) \citet{bissell-etal-2025-theoretical}\newline
\\
\hline
Narrative Empathy\newline\cite{keen2006theory} & \checkmark & \checkmark & \checkmark & \checkmark & Literary\newline Studies &
(U) \citet{shen-etal-2023-modeling,shen_heart-felt_2024}\newline
(U) \citet{liu2025raemollm}
\\
\hline
Basic Elements of Narrative\newline\citep{herman2009basic} & \checkmark & \checkmark & \checkmark & \checkmark & \textbf{Postclassical\newline Narratology} &
(U) \citet{antoniak-etal-2024-people}\newline
(U) \citet{sui_confabulation_2024}\newline
(U) \citet{piper_narradetect_2025}
\\
\hline
Migration Studies\newline\citep{dennison2021narratives,komendantova2023misinformation}\newline Media Studies\newline\citep{marino2024polarization}\newline Communication Theory\newline\citep{hameleers2023disinformation} & \checkmark & \checkmark & \checkmark & \checkmark & \textbf{Social\newline Sciences} &
(U) \citet{nikolaidis-etal-2025-polynarrative}\newline
(G) \citet{vykopal2024disinformation}\newline
(U) \citet{maggini2026partisanlens}\newline
(U) \citet{schneider2026llm}\newline

\\




\bottomrule
\end{tabular}
\caption{Narrative Theories. Level, origins, and application examples across 
understanding (U), generation (G), or both (B) tasks.}
\label{tab:NT_table}
\end{table}


\subsection{\redtext{From Aristotle to Propp}}

Before the emergence of classical narratology as a discipline in the 20th century, the study of formal textual features that are common between all narratives can trace its lineage from \citet{aristotle_poetics_penguin} Poetics, \citet{freytag1894technik} dramatic pyramid, Russian formalist distinctions to \citet{propp1928morfologiya,propp1968morphology} morphology of the folktale. Although these early theories differ in historical origin, they decompose narrative into structural units. In LLM-based methods, these theories most often operate at the fabula and discourse levels. They define what kinds of events should occur, how those events should be ordered, and how narrative tension, conflict, or transformation should develop across a story.

\cite{aristotle_poetics_penguin}'s \textit{Poetics} is one of the earliest sources for treating narrative as a composition of smaller distinguishable elements. \citet{mirowski2023co}'s Dramatron is a human-in-the-loop hierarchical co-generation system for theatre scripts and screenplays. Rather than prompting an LLM to generate a complete script in a single pass, Dramatron decomposes scriptwriting into linked stages: premise or theme, characters, plot beats, locations, and dialogue. This hierarchical prompt strategy loosely echoes Aristotelian categories such as \textit{dianoia} (thought or theme), \textit{mythos} (plot), \textit{ethos} (character), and \textit{lexis} (diction or dialogue). Dramatron is an example of an approach where high-level dramatic structure is planned first, and lower-level discourse realisation is generated afterwards.

\citet{song2024conflict}'s CNGCI, a conflict-driven narrative generation framework, applies \citet{freytag1894technik}'s dramatic pyramid, which models narrative progression as the introduction of an inciting incident that drives rising tension toward a climax. CNGCI's framework defines ``conflict'' as a logical obstacle that weakens the expected causal link between a protagonist's initial context and their objective. The system uses a multi-step pipeline. First, the protagonist's goal is inferred from the context describe in an input sentence, using a neuro-symbolic method which pairs a knowledge graph with a transformer \citep{hwang2021comet}. To introduce a conflict that drives rising tension, the system uses Defeasible Natural Language Inference ($\delta$-NLI) \citep{rudinger-etal-2020-thinking} to generate an obstacle that reduces the probability of the goal occurring given the context, satisfying the condition $P(\text{Goal} \mid \text{Context}, \text{Obstacle}) < P(\text{Goal} \mid \text{Context})$. Finally, this tuple (context, goal, and generated obstacle) is passed to a fine-tuned GPT-2 language model, which transforms the input tuple into a coherent short story.

The distinction between fabula and sujet/syuzhet from Russian formalism has been used by graph-based and multi-agent systems (MAS) to separate story-world simulation from narrative presentation. \citet{yoo2024leveraging} use Story Plan Graphs to select a single sequence narrative events from different possible branches. Meanwhile, \citet{yu-etal-2025-multi}'s MAS uses a director agent to select and command character agents to respond, the conversation history is then given to a rewrite agent which then outputs the final story in accordance to the specified presentation outline (which may be non-chronological). In both cases, the formalist distinction informs the design of components in compound systems: one component models causal or chronological events, while another determines how the story is presented to the reader.

\citet{propp1928morfologiya,propp1968morphology}'s morphology offers a structural model which includes universal patterns of character or story functions. Because Propp's functions are discrete and label-like, they are especially applicable for annotation, classification, and controlled generation. \citet{gervas2024storytelling} tests whether LLMs can annotate folktale synopses directly with Proppian character functions in zero-shot and few-shot settings. Proppian morphology also continues to influence story generation in systems that rely on goal-driven or function-based fairy-tale structure. \citet{ermolaeva-etal-2024-tame}, for instance, develop an interactive fairy-tale generation framework that uses story goals, emotional arcs, and prompt-controlled stage transitions. While not every system applies Propp's morphology in full, these systems inherit the structuralist idea that stories can be guided by abstract functional stages.

\subsection{Classical Narratology}


Classical Narratology continues to seek universal narrative structures. \cite{todorov19712}'s theory of narrative equilibrium which describes the transformations of one narrative state to the next has given influence to the formulation of narrative as a system of causality used by \citep{heddaya-etal-2024-causal,ghaffari-hokamp-2025-narrative,das-etal-2024-media,liu2026retell}. For example, \cite{ghaffari-hokamp-2025-narrative} applies this definition to introduce a framework for navigating stories as dynamic causal event graphs\bluetext{, while \citet{liu2026retell} synthesise Todorov's theory with the post-classical notion of narrativity to define reward scores used by reinforcement learning with AI feedback for logical, rational and structurally complete counterfactual story generation.} 

In another structuralist theory, \citet{barthes1975introduction} defines cardinal functions (or nuclei) as the essential, irreplaceable plot components of a narrative. While \citet{wilmot2021memory} applies the theory with LLMs to label the per-sentence salience of novel/play summaries, \citet{huang-usbeck-2024-narration} applies the same theory to extract event causal relations from non-fictional news stories.

Classical Narratology continues the Russian Formalist tradition of separating fabula/sujet: historie/recit \citep{genette1980narrative} and story/discourse \citep{chatman1980story}. This separation is used by \citet{shen2025adapting} to distinguish between story tasks (actions, characters, settings) and discourse tasks (turning point, summarisation) for narrative understanding tasks in the movie domain. Similarly, the intentional separation of these narrative levels is also implemented by \cite{chen-si-2024-reflections} and \cite{halperin2024artificial} in story generation tasks, and \cite{hatzel-biemann-2024-story} when creating embeddings of narrative texts using last-token representations. \citet{zhao2023parrot}'s PARROT training method pairs two parallel narratives that tell the same underlying story, using one discourse as a supervision signal to guide comprehension of the other discourse. Trough this process, the model is forced to learn to model the common fabula, achieving comparable performance to fully supervised models on narrative comprehension benchmarks without any task-specific training data.

In addition to the fabula/discourse levels, Classical Narratology also examines the function of the narration level. For feature labelling, \cite{piper-bagga-2024-using} operationalise \cite{genette1980narrative}’s discourse theory of the fabula/discourse/narration triangle by using LLMs to generate labels (validated by human annotators) that correspond to narrative features such as mood, voice and tense (e.g., pastness, anachrony, POV).

\cite{booth1961rhetoric} and \cite{chatman1980story}'s theories both explore the relationship between the narrator and the intentions of implied author. Studies apply these theories to investigate: how well LLMs can detect different types of unreliable narrators using textual features such as tone, inconsistencies and digressions \citep{brei-etal-2025-classifying}, how well instruction-tuned LLMs can simulate meaningfully different authorial voices when narrating the same events \citep{hamilton-2024-detecting}, labelling the morals of folktales, novels, movies/TV, personal stories from social media and the news \citep{hobson-etal-2024-story}, while \cite{zhou-etal-2024-large} specialises moral labelling of climate news.



\subsection{Postclassical Narratology}

Here we show applications of postclassical narratology which expands beyond classical formal/structural boundaries. Specifically, these approaches conceptualise narrativity as a scalar property and studies the phenomena of narrative beyond the textual artifact. By integrating frameworks from cognitive science, media theory, and cultural studies, postclassical narratology is an interdisciplinary approach to theories of narrative.

\subsubsection{\bluetext{Scalar Narrativity}}

An important change from classical to postclassical narratology is the shift from defining narrativity as a binary class to a scalar quantity \citep{herman2009basic,piper2021narrative}. Narrativity denotes the extent to which a text or discourse exhibits the qualities of narrative, including temporal progression, causal linkage, and the presence of agents and events. For example, a chronologically ordered news report may possess lower narrativity than a novel with rich character development and plot dynamics. \cite{piper2021narrative} annotate passages based on narrativity scores that reflect the degree to which the text engages with the act of narration, characterised by the dimensions: agency, event sequencing and world building. These annotated passages are used to test LLMs’ detection of narrative features, thus measuring how well models capture and distinguish narrative from non-narrative textual qualities and their prediction of a scalar value \citep{piper_narradetect_2025}. \cite{sui_confabulation_2024} use \citet{piper2021narrative}'s formulation of narrativity to compare LLM confabulations (hallucinations) with factually accurate responses. They find that LLM confabulations display higher narrativity, and make the novel argument that confabulations mirror human cognitive sense-making and storytelling tendencies \redtext{because of narrative's role in enabling coherent, articulate and consistent forms of communication.} \bluetext{\cite{steg-etal-2022-computational} and \bluetext{\cite{liu2026retell} also annotate passages using scalar narrativity values based on definitions from \cite{sternberg1990telling} and \cite{todorov19712} respectively.}}

Furthermore, \cite{mire-etal-2024-empirical} investigate crowd workers' and LLMs' perceptions of narrativity to explore the variability of narrative perceptions without giving annotators singular definitions of narratives. Comparisons find that GPT-4 is less likely to identify stories in texts that describe abstract actions (e.g., habits, behaviours, processes) compared to human crowd workers. he study finds that the crowd workers' perceptions of narrative are based primarily on a few core textual features (events, characters, plot) while extra-textual features, such as cognitive and aesthetic experiences while reading (sense of conflict, cohesion, feeling like a story), are also important. The findings are largely consistent with prior work and provide empirical validation for prior theories \citep{herman2009basic,piper2021narrative}. Understanding alignment between LLM and human annotators when it comes to narrative understanding is crucial for future work that use LLMs to approximate human annotators. 

\subsubsection{Cognitive \redtext{Narrative Theories}}

\cite{keen2006theory}'s theory of narrative empathy theorises that narratives have unique capacities to evoke empathy in comparison to other forms of non-narrative communication. \cite{shen-etal-2023-modeling} apply this to model empathic similarity between story pairs, operationalising resonance through events, emotional trajectories, and morals, with a fine-tuned retrieval model outperforming semantic similarity baselines in a 150-person study. \cite{shen_heart-felt_2024} further investigates which narrative style features causally elicit empathy, finding through LLM-annotated HEART taxonomy features that vividness of emotion and plot volume are the primary pathways. \bluetext{\cite{liu2025raemollm} references \cite{keen2006theory} to justify their framework which uses sentiment and emotion features to effectively classify content as misinformation across topical domains without fine-tuning, achieving improvements of 15.64\%, 31.18\%, and 15.73\% over few-shot baselines on fake news, rumour, and conspiracy theory detection respectively.}

\bluetext{Similarly focusing on emotional responses, \cite{sternberg1990telling}'s theory of narrative universals propose that narrativity can be define by the reader's experience of suspense, curiosity and surprise. This theory is applied by \cite{steg-etal-2022-computational} in a human annotation experiment, and \cite{bissell-etal-2025-theoretical}'s proposed framework to specifically measure surprise in human and LLM authored story endings--finding that human readers preferred more surprising stories written by human authors.}

Reader-centred approaches inform the evaluation of LLM-generated stories when using human evaluation and LLM-as-judge.  \cite{harel-canada-etal-2024-measuring} employ the Psychological Depth Scale, grounded in cognitive frameworks like reader-response criticism \citep{holland1989dynamics} and text world theory \citep{gavins2007text} to assess stories on dimensions such as emotion provocation, empathy, engagement, authenticity, and narrative complexity: factors which shift attention from narrative structure alone to the reader's cognitive and affective construction of a story world.



\bluetext{\cite{gupta-etal-2025-automated} present an application of using LLMs for narrative understanding in clinical settings, where story retelling is used to assess cognitive communication abilities in aphasia\footnote{A language impairment caused by stroke or other brain injury that affects a person’s ability to express and understand written and spoken language \citep{gupta-etal-2025-automated}.} patients. Likewise, work on personal or therapeutic narratives, such as \citet{ma2026multi} and \citet{kim-etal-2025-share-story}, uses narrative as a means of analysing self-expression and metacognition. In these cases, narrative is studied as an interactional and psychological practice rather than only as a textual structure.} \cite{sun-etal-2024-event} reference psychological theories of narrative as a sense-making cognitive faculty for causality\footnote{The claim of narrative as a cognitive system of causality is contradicted by \cite{bruner1991narrative}, whose psychological narrative theory distinguishes between the Paradigmatic or Logico-Scientific Mode of causal explanations and the narrative mode which focuses on beliefs, values and desires. We emphasise that the landscape of narrative studies is full of contradictions and lacks unified consensus.} \citep{trabasso1985causal,keefe1993time,graesser2003does}. Cognitive capabilities such as theory-of-mind are tested through narrative tasks in LLMs \citep{zhou-etal-2025-essence}., and \cite{de-langis-etal-2025-llms} use narratives to evaluate LLM processing of temporal meaning, finding persistent struggles with causal reasoning and narrative comprehension.

\subsubsection{Contextual \redtext{Narrative Theories}}




\redtext{Researchers examine how narratives are produced, circulated, and mobilised across specific social, cultural, political, and institutional contexts. They apply narrative theories to non-literary texts--such as AI generated narratives, news articles and social media posts. In these cases, LLMs enable the large-scale analysis of narrative as a social phenomenon.}

Several papers in our survey fall into this broader category, treating narrative as evidence of social values, collective identities, and cultural assumptions. For example, \cite{beguvs2024experimental} explore cultural paradigms by examining how human- and machine-generated stories reflect the Pygmalion myth.\footnote{An ancient Greek myth about a sculptor who falls in love with his human-like creation: an ivory statue of a woman who comes to life. A modern retelling of the Pygmalion myth can be found in the film \textit{Ruby Sparks} \citep{ruby_sparks_2012}.} Similarly, \citet{zhou-etal-2024-large} use LLMs to analyse climate-change news across regions, identifying roles, moral framings, and patterns of agency to show how the topic is narrativised in situated media discourse; \citet{hamilton2022covid} apply a training-based approach to pandemic news: fine-tuning a pre-pandemic GPT-2 on CBC News to generate a counterfactual corpus of what COVID-naive coverage might have looked like, then comparing it against actual pandemic reporting to surface distinctive human narrative framing choices.

Social sciences' studies of migration and policy discourse, including work by \citet{dennison2021narratives}, \citet{komendantova2023misinformation}, and \citet{otmakhova-frermann-2025-narrative}, establish frameworks to investigate how public narratives shape perceptions of social issues. Additionally, works by \citet{nikolaidis-etal-2025-polynarrative}, \citet{piskorski-etal-2025-semeval}, \citet{maggini2026partisanlens}, and \citet{schneider2026llm} use LLM-assisted annotation to classify news, political communication, and manipulative messaging. These papers treat storytelling as a mechanism through which public issues are communicated and contested, demonstrating how narrative categories can be operationalised at scale for media analysis.

\subsection{\redtext{Other Theories}}
\label{subsubsec:other_narratology}

\redtext{We additionally explore applications of other narrative theories not easily categorised by the narratology lineage, including: sociolinguistics, theories from film and pop culture, and distant reading from literary studies.}

\subsubsection{\redtext{Sociolinguistics}}

\cite{labov1972language}'s social linguistic theory is used by \cite{troiano_clause-atlas_2024} to annotate novels by classifying clauses as event, subjective experiences and contextual clauses, then using the output distribution to quantitatively analyse the narrative structures. \citet{watahiki2026developing} extend Labovian narrative structure analysis to Japanese oral narratives, developing annotation guidelines that adapt the sociolinguistic framework to a non-Western language context.

While it has been mentioned above that \cite{shen_heart-felt_2024} draws from cognitive narrative theories, they also draw on \cite{van2017evoking}'s framework of linguistic cues such as emotional/cognitive subjects, character development, and perception to facilitate character identification and immersive experiences. NLP researchers have combined and synthesised theories from different fields in their research.

\subsubsection{\redtext{Pop Culture Narrative Theories}}

Several structural narrative theories have emerged from books, film studies and screenwriting. We refer to these as pop culture narrative theories due to their popularity in our culture beyond academic contexts.

\citet{tian-etal-2024-large-language} directly cites and applies \citet{vonnegut_shapes_of_stories}'s Shape of Stories in story generation through a planning and prompting method: story progression is represented through changing affective valence over time. In contrast to Proppian functions or Campbellian stages, story-shape approaches do not require labelling individual events with narrative roles. Instead, they model the global emotional contour of a narrative, making them especially relevant to evaluation, comparison, and literary analysis. \citet{mori2022plug} apply the same principle to story continuation where their emotion-aware framework uses Vonnegut-derived affective arc targets to condition story completion. Other example applications of the theory illustrate how narrative theories often enter NLP through intermediaries. Rather than citing \citet{vonnegut_shapes_of_stories} directly, computational work may cite inspired applications of story shape through sentiment arcs or emotional trajectories--as in \citet{bizzoni-feldkamp-2023-comparing}, who cite \citet{reagan2016emotional}'s sentiment arc framework, which itself draws on \citet{vonnegut_shapes_of_stories}, demonstrating how theoretical narrative concepts can enter NLP indirectly.

\cite{chun_aistorysimilarity_2024} references concepts from narratology broadly and also theories from screenwriting \citet{snyder2005save,mckee1997substance,truby2008anatomy} for its four dimensions of storytelling (themes, setting, plot and characters). Screenwriting \citep{mckee1997substance} is also referenced by \citet{ran-etal-2025-bookworld}, which uses `scene' as the `minimal narrative unit'.

\cite{Campbell1949Hero,campbell2008hero}'s monomyth and related hero's journey frameworks are used in LLM generation methods as templates for planning. \citet{wang2025generating}'s DOME, Dynamic hierarchical Outlining with Memory-Enhancement, combines a dynamic hierarchical outline with a temporal knowledge graph memory module: the outline supports macro-level plot planning for long-form story generation, while the temporal knowledge graph stores and retrieves generated content to reduce contextual contradictions, incorporating Campbell-inspired staged progression into the rough-outline generation process. \citet{lima2024pattern} provide a more explicitly pattern-oriented use of the monomyth. Their PatternTeller system uses narrative patterns, including the Hero's Journey, to guide LLM-based story composition, and applies the most specific generalisation (MSG) criterion to identify new reusable narrative structures from examples.

\subsubsection{Distant Reading}

Distant reading is an approach in literary studies associated with the scholar Franco Moretti \citep{moretti2005graphs}. \cite{hamilton-etal-2025-city} explore concepts from network theory \citep{moretti2011network} by extracting social networks from multilingual fiction and non-fiction narratives to construct structured datasets of story worlds. In the age of LLMs, distant reading offers a powerful approach to analysing narratives at scale. Similarly, \cite{bizzoni-feldkamp-2023-comparing} track sentiment variance, \citet{yang-anderson-2024-evaluating} track character similarity and \cite{sancheti-rudinger-2025-tracking} track character relationships across entire novels, demonstrating how computational tools can capture narrative dynamics at scale.

\subsection{\redtext{Theoretical Contributions}}
\label{subsec:theoretical_contributions}

\redtext{Theoretical contribution in this survey takes three forms: synthesis, taxonomy, and validation. Synthesis combines existing frameworks to produce new analytical tools; taxonomy operationalises theoretical concepts into measurable categories; and validation uses computational results to test whether theoretical claims hold empirically.}

\subsubsection{\redtext{Synthesis and Taxonomy}}

In terms of synthesis and taxonomy, researchers combine existing narrative frameworks to evaluate or generate text. \citet{piper-bagga-2024-using} build on \citet{genette1980narrative}'s tripartite framework of tense, mood, and voice, then update it with \citet{herman2009basic}'s postclassical definitions of sequentiality, world-building, and qualia to derive fifteen measurable discourse-level features. \citet{tian-etal-2024-large-language} draw on both \citet{vonnegut_shapes_of_stories} and \citet{vanDijk_macrostructures}, theories from entirely different traditions, to model narrative structure as a progression of affective valence. \citet{liu2026retell} use synthesis to define a reward function: they taxonomise \citet{todorov19712}'s narrative equilibrium and combine the detection of structural stages with the postclassical notion of scalar narrativity as reward scores for reinforcement learning, making narrative theory directly applicable to generating a training signal.

Cross-disciplinary syntheses are also notable. The HEART framework \citep{shen_heart-felt_2024} combines \citet{keen2006theory}'s theory of narrative empathy with \citet{van2017evoking}'s sociolinguistic framework of linguistic identification cues to define a taxonomy of narrative elements that elicit empathetic responses. \citet{harel-canada-etal-2024-measuring} synthesise reader-response criticism and cognitive narrative theory into the Psychological Depth Scale to evaluate narratives. \citet{sterner-etal-2026-contrastive} apply \citet{labov1967narrative}'s principle that salient narrative events are those which advance a narrative rather than merely describe it, training story embeddings on pairs sharing the same plot but differing in surface form, and showing that this theoretically grounded objective outperforms masked-language-model baselines. 


\subsubsection{\redtext{Empirical Validation}}

\redtext{A recurring contribution in the surveyed work is the empirical validation of narrative theories through their operationalisation into NLP frameworks. Translating qualitative concepts into explicit, testable criteria generates large-scale empirical findings that the theory alone could not produce.}

\citet{bissell-etal-2025-theoretical} adapt \citet{sternberg1990telling}'s theory of narrative surprise into six annotation criteria: initiatoriness, immutability violation, predictability, post-dictability, importance, and valence; they find that only four correlate significantly with reader preferences. \citet{brei-etal-2025-classifying} establish a three-level taxonomy of narrator unreliability--intra-narrational, inter-narrational, and inter-textual--extending \citet{booth1961rhetoric}'s concept into a categories for annotation. The finding that inter-textual unreliability is significantly harder to detect than intra-narrational unreliability implies that the most demanding form of the phenomenon depends on knowledge that exists outside the text itself, which is a postclassical rather than classical understanding of the narrator. \citet{mitran2025probing} develop the Moral Foundations Character Action Questionnaire (MFCAQ), a character-centric framework applying \cite{haidt2004intuitive}'s Moral Foundations Theory to narrative, and apply it to 2,697 folktales from 55 countries. They find broad cross-cultural distribution of moral foundations with significant consistency across regions, and a more balanced distribution of positive and negative moral content than suggested by prior work--extending moral schema analysis beyond Western literary contexts.


\citet{hamilton-etal-2025-city} apply \citet{moretti2011network}'s network theory through large-scale distant reading across multilingual fiction and non-fiction: providing empirical evidence that social networks in fiction are smaller and more tightly clustered; non-fiction networks show greater community diversity. \citet{beguvs2024experimental} validate cultural narrative claims by comparing human and LLM-generated stories centred on the Pygmalion myth, finding that LLMs reproduce and in some cases amplify canonical gender dynamics, providing large-scale empirical grounding for claims about the persistence of cultural schemas in narrative roles and structures. \citet{bissell-etal-2025-theoretical} find that human-authored story endings consistently outperform LLM-authored endings on narrative surprise and preference, validating \citet{sternberg1990telling}'s claim that the experience of surprise is fundamental to the narrative experience. \citet{wilmot2021memory} integrate annotated Proppian functions with \citet{barthes1975introduction}'s theory of cardinal functions to evaluate LLM identification of salient events, implicitly testing whether two independently developed structural theories converge on the same notion of narrative importance. \citet{liu2026retell} test whether Todorov formal narrative equilibrium features are formally enough to measure narrativity and predict correlation with human preferences.

\citet{mire-etal-2024-empirical} find that crowd workers' narrativity judgements depend primarily on textual features (events, characters, plot) consistent with existing theory \citep{piper2021narrative}, but are additionally shaped by extra-textual cognitive and aesthetic responses--such as a sense of cohesion or ``feeling like a story''--that theory-based definitions do not capture, challenging the classical structuralist assumption that textual features alone are sufficient for narrative detection and providing empirical support for the postclassical notion of a narrative's cognitive and contextual situatedness.

Another novel empirical contribution in the survey comes from \citet{sui_confabulation_2024}, who apply \citet{piper2021narrative}'s formulation of narrativity to compare LLM confabulations with factually accurate responses. Finding that hallucinated outputs display significantly higher narrativity, they argue that confabulations in LLMs mirror human cognitive sense-making because narrative is the mind's default mode for producing coherent causal accounts. This claim was not predicted by existing narrative theory--which lacked access to the kind of large-scale controlled comparison LLMs make possible--and it repositions the LLM not as a tool for studying narratives but as a theoretical object in itself. If a model's errors are more narrative than its correct outputs, narrative structure is not merely a feature of training data but is actively reinforced by the objectives and architectures through which these models learn.



\section{Datasets and Tasks}\label{sec_data_task} 

This section examines the narrative datasets and tasks explored in surveyed LLM-based narrative research. In the opening sentence of ``An Introduction to the Structural Analysis of Narrative'', \citet{barthes1975introduction} noted: ``There are countless forms of narrative in the world.'' \citet{moretti2000_Slaughterhouse} argues that literary studies, relying on close reading, attends only to canonical texts while ignoring 99.5\% of existing literature--texts he describes as the `great unread', lost to the `slaughterhouse of literature'. Against this backdrop, \citet{piper2021narrative} identify three core challenges for narrative studies at the intersection of NLP and the humanities: economies (how real-world contexts shape narrative production), responses (how textual features influence reader, viewer, or listener reactions), and beliefs (how individual and social narratives underpin human behaviour). Addressing these challenges requires studying diverse narratives at scale, a need directly reflected in the datasets surveyed here, which span a wide range of narrative forms and tasks.

\redtext{\citet{hamilton-etal-2026-narrabench}'s Narrabench provide a useful survey showing which datasets can be used for different types of narrative understanding benchmarks, and \citet{zhu-etal-2023-nlp}'s survey  also details narrative understanding datasets and tasks that readers may find useful. \citet{piper-2023-computational} identified multilingual modelling as a core challenge for the field, calling for deeper investment in cross-cultural and multimodal narrative datasets. The vast majority of surveyed work uses English-only data, and the theoretical frameworks remain rooted in Western narrative traditions.}

{\footnotesize
\setlength{\tabcolsep}{1pt} 
\begin{xltabular}{\textwidth}{>{\raggedright\arraybackslash}p{1.5cm} >{\raggedright\arraybackslash}p{4cm} c c c c X} 
    \caption{Dataset Overview, Sources and Usage}
    \label{tab:dataset_overview} \\
    \toprule
    \multirow{2}{*}{\textbf{Source}} & 
    \multirow{2}{*}{\textbf{Included in Dataset}} & 
    \multicolumn{4}{c}{\textbf{Source Type}} & 
    \multirow{2}{*}{\textbf{Usage}} \\
    
    \cmidrule(lr){3-6}
    
    & & 
    \rotatebox{90}{Fiction} & 
    \rotatebox{90}{Non-fiction} & 
    \rotatebox{90}{Personal} & 
    \rotatebox{90}{Elicited} & \\
    \midrule
    \endfirsthead

    \caption[]{Dataset Overview, Sources and Usage (Continued)} \\
    \toprule
    \multirow{2}{*}{\textbf{Source}} & 
    \multirow{2}{*}{\textbf{Included in Dataset}} & 
    \multicolumn{4}{c}{\textbf{Source Type}} & 
    \multirow{2}{*}{\textbf{Usage}} \\
    
    \cmidrule(lr){3-6}
    
    & & 
    \rotatebox{90}{Fiction} & 
    \rotatebox{90}{Non-fiction} & 
    \rotatebox{90}{Personal} & 
    \rotatebox{90}{Elicited} & \\
    \midrule
    \endhead

    \midrule
    \multicolumn{7}{r}{\textit{Continued on next page}} \\
    \endfoot

    \bottomrule
    \endlastfoot

    Reddit & 
    AskReddit Turning Points \citep{ouyang2015modeling} & 
    & & \checkmark & & 
    Moral Identification \citep{hobson-etal-2024-story}; Discourse Feature Identification \citep{piper-bagga-2024-using} \\

    & 
    StorySeeker \citep{antoniak-etal-2024-people} & 
    & & \checkmark & & 
    Narrative Detection or Extraction \citep{mire-etal-2024-empirical,piper_narradetect_2025} \\

    & 
    WritingPrompts \citep{fan-etal-2018-hierarchical} & 
    \checkmark & & & & 
    Story Continuation \citep{xie-etal-2023-next}; Story Quality Criteria \citep{harel-canada-etal-2024-measuring} \\

    & 
    r/aimitheasshole & 
    & & \checkmark & & 
    Discourse Feature Identification \citep{brei-etal-2025-classifying} \\
    
    \midrule 
    
    Books & 
    Project Gutenberg & 
    \checkmark & \checkmark & & & 
    Embedding \citep{hatzel-biemann-2024-story}; Discourse Feature Identification \citep{brei-etal-2025-classifying}; Character Relationship Identification \citep{sancheti-rudinger-2025-tracking} \\
    & 
    Novels & 
    \checkmark & & & & 
    Character Role Identification \citep{yang-anderson-2024-evaluating}; Fabula Simulation/Representation \citep{ran-etal-2025-bookworld}; Sentiment Analysis \citep{bizzoni-feldkamp-2023-comparing}; Topic Labelling \citep{piper-wu-2025-evaluating}; Story Quality Criteria \citep{yang-jin-2025-matters} \\
    & 
    CONLIT \citep{piper2022conlit} & 
    \checkmark & \checkmark & & & 
    Character Relationship Identification \citep{piper-etal-2024-social} \\

    \midrule 
    
    News & 
    Media Frames Corpus \citep{card-etal-2015-media} & 
    & \checkmark & & & 
    Narrative Detection or Extraction \citep{das-etal-2024-media} \\
    & 
    Now Corpus \citep{davies2016now} & 
    & \checkmark & & & 
    Narrative Detection or Extraction \citep{heddaya-etal-2024-causal} \\
    & 
    Meta URL Shares & 
    & \checkmark & & & 
    Narrative Detection or Extraction \citep{marino2024integrating} \\
    & 
    PolyNarrative \citep{nikolaidis-etal-2025-polynarrative} & 
    & \checkmark & & &
    Narrative Detection or Extraction \citep{piskorski-etal-2025-semeval}; Character Role Identification \citep{piskorski-etal-2025-semeval} \\
    & 
    CNN News \citep{hermann2015teaching} & 
    & \checkmark & & &
    Story Continuation \citep{xie-etal-2023-next} \\
     & 
    Climate News & 
    & \checkmark & & &
    Moral Identification \citep{zhou-etal-2024-large} \\
    & 
    FakeNewsAMT and Celebrity \citep{perez2018automatic} & 
    & \checkmark & & &
    Misinformation Detection \citep{liu2025raemollm} \\
    & 
    Global News \citep{kumar_saksham_2023} & 
    & \checkmark & & &
    Moral Identification \citep{hobson-etal-2024-story}; Topic Labelling \citep{piper-wu-2025-evaluating} \\
    & 
    PartisanLens \citep{maggini2026partisanlens} & 
    & \checkmark & & &
    Misinformation Detection \citep{maggini2026partisanlens}\\
    & 
    COVID-19 News & 
    & \checkmark & & &
    Controlled Story Generation \citep{hamilton2022covid}\\

    \midrule 

    ROCStories & 
    ROCStories \citep{mostafazadeh2016corpus} & 
    & & & \checkmark & 
    Controlled Story Generation \citep{gangal2022nareor,brei2024returning,song2024conflict}; Story Continuation \citep{mori2022plug}; Embedding \citep{hatzel-biemann-2024-story,sterner-etal-2026-contrastive} \\

    & 
    Story Commonsense \citep{rashkin2018modeling} & 
    & & & \checkmark & 
    Controlled Story Generation \citep{xie2022psychology} \\

    & 
    TimeTravel \citep{qin-etal-2019-counterfactual} & 
    & & & \checkmark & 
    Story Continuation \citep{liu2026retell} \\

    & 
    COPES \citep{wang-etal-2023-cola} & 
    & & & \checkmark & 
    Event Segmentation \citep{sun-etal-2024-event} \\
    
    \midrule 
    
    Crowd Workers & 
    HIPPOCORPUS \citep{sap2020recollection} & 
    \checkmark &  & \checkmark & \checkmark & 
    Empathy Detection \citep{shen-etal-2023-modeling,shen_heart-felt_2024} \\

    & 
    Tell Me A Story \citep{huot2024agents} & 
    \checkmark &  &  & \checkmark & 
    Controlled Story Generation \citep{yu-etal-2025-multi} \\
    
    \midrule 

    Wikipedia & 
    Tell Me Again! \citep{hatzel2024tell} & 
    \checkmark & \checkmark & & \checkmark & 
    Embedding \citep{hatzel-biemann-2024-story} \\
    & 
    Movie Remakes Plots \citep{chaturvedi2018have} & 
    \checkmark & & & & 
    Embedding \citep{hatzel-biemann-2024-story} \\
    & 
    Movie Synopses & 
    \checkmark & & & & 
    Controlled Story Generation \citep{tian-etal-2024-large-language} \\

    \midrule 

    Film/TV & 
    Screenplays & 
    \checkmark & & & & 
    Embedding \citep{chun_aistorysimilarity_2024} \\
    & 
    NarraSum \citep{zhao2022narrasum} & 
    \checkmark & & & & 
    Question Answering \citep{zhao2022narrasum}; Moral Identification \citep{hobson-etal-2024-story} \\
    & 
    MovieGraph \citep{vicol2018moviegraphs}, MovieNet \citep{huang2020movienet} & 
    \checkmark & & & & 
    Narrative Understanding (Mixed) \citep{shen2025adapting} \\
    & 
    MovieGraph \citep{papalampidi2019movie} & 
    \checkmark & & & & 
    Embedding \citep{sterner-etal-2026-contrastive} \\

    \midrule 

    Plays & 
    Shmoop Corpus \citep{chaudhury2019shmoop} & 
    \checkmark & & & & 
    Event Segmentation \citep{wilmot2021memory} \\
    & 
    DraCor \citep{fischer2019programmable} & 
    \checkmark & & & & 
    Character Relationship Identification \citep{udrea2025speaks} \\

    \midrule 

    Folktales & 
    Unspecified & 
    \checkmark & & & & 
    Character Role Identification \citep{gervas2024tagging} \\
    & 
    Kaggle Folk Tales Dataset & 
    \checkmark & & & & 
    Moral Identification \citep{hobson-etal-2024-story,mitran2025probing} \\
    & 
    Aesop's Fables Dataset \citep{aesop_fables,ardiputra2021aesop} & 
    \checkmark & & & & 
    Summarisation \citep{chen-si-2024-reflections} \\
    & 
    Propp Dataset \citep{finlayson2017propplearner} & 
    \checkmark & & & & 
    Event Segmentation \citep{wilmot2021memory} \\

    \midrule 

    Podcasts & 
    Spotify Podcast Dataset \citep{clifton2020100} & 
    \checkmark & \checkmark & \checkmark & & 
    Narrative Detection or Extraction \citep{abdessamed2024identifying} \\
    
    \midrule 

    LLM Generations & 
    FaithDial \citep{dziri-etal-2022-faithdial}, BEGIN \citep{dziri-etal-2022-evaluating}, HaluEval \citep{li-etal-2023-halueval} & 
    & & & & 
    Narrative Detection or Extraction \citep{sui_confabulation_2024} \\
    & 
    StoryWriter \citep{xia2025storywriter} & 
    & & & & 
    \\

    \midrule 

    Blogs/Reviews & 
    PersonaBank \citep{lukin2016personabank} & 
    & & & \checkmark & 
    Discourse Feature Identification \citep{brei-etal-2025-classifying} \\

    & 
    Deceptive Reviews \citep{ott-etal-2013-negative} & 
    & & & \checkmark & 
    Discourse Feature Identification \citep{brei-etal-2025-classifying} \\

    \midrule 

    Twitter & 
    PHEME \citep{zubiaga2017exploiting}, COCO \citep{langguth2023coco} & 
    & \checkmark & & & 
    Misinformation Detection \citep{liu2025raemollm} \\

    \midrule 

    Therapy Narratives & 
    Chinese Therapeutic Writing Samples \citep{ma2026multi} & 
    & &\checkmark & & 
    Mental Health Prediction \citep{ma2026multi} \\
    
    \midrule 
    
    Mixed Sources & 
    Empathetic Stories \citep{shen-etal-2023-modeling} & 
    & & \checkmark & \checkmark & 
    Discourse Feature Identification \citep{shen_heart-felt_2024} \\

    & 
    NarraDetect \citep{piper2022toward} & 
    \checkmark & \checkmark & \checkmark & & 
    Narrative Detection or Extraction \citep{piper_narradetect_2025} \\

\end{xltabular}
}

\subsection{\redtext{Types of Narrative Sources}}
\label{sec:narrative_sources}

For tasks related to narrative understanding and generation, NLP researchers draw on textual data from diverse sources, including fictional literary narratives, non-fictional narratives, personal narratives, and elicited narratives (Table~\ref{tab:dataset_overview}). Different sources suit different aims, for example: cultural investigations may use news articles or folktales \citep{hobson-etal-2024-story}; model performance-focused studies often use short stories for their brevity and diversity \citep{chen-si-2024-reflections}; and literary traditions of distant reading \citep{moretti2005graphs} typically rely on novels and \redtext{long-form} texts \citep{hamilton-etal-2025-city}.

Key challenges in using narrative data include defining annotation schemes, extracting narratives from unstructured data and varied document types (e.g., social media\footnote{\cite{page2018narratives} offers an insightful account of social media within the sociolinguistic paradigm of `small stories'}), and developing standard evaluation frameworks comprising datasets, baselines, and metrics \citep{santana2023survey}. Many narrative datasets require expert annotation \citep{shen_heart-felt_2024,piper_narradetect_2025}. Large-scale annotation typically involves expert annotators labelling a subset as ground truth, followed by less-experienced annotators labelling the remainder. Annotation validity is assessed via agreement scores and by evaluating model performance (using NLP metrics such as accuracy, precision, recall, and F1) on labelled data \citep{piper-etal-2024-social}.

Researchers rarely explain why a given corpus qualifies as narrative, typically assuming the category is self-evident--for forms such as short stories, novels, and news articles. The need to explicitly establish a definition of narrative and selection criteria exists for datasets that contain both narratives and non-narratives for narrative detection \citep{antoniak-etal-2024-people, piper_narradetect_2025}. These datasets draw from literary narrative theory to define a taxonomy of textual features that contribute to narrativity from the works of \citet{bal2009narratology}, \citet{prince2003dictionary}, and \citet{herman2009basic}. \redtext{\citet{piper_narradetect_2025}, for instance, use three human annotators giving passage-level ratings on the criteria of agency, event sequencing, and world building \citep{herman2009basic} to classify the scalar narrativity of 400 texts drawn from genres as varied as memoirs, academic articles and book reviews.}

We use four categories to identify different narrative sources. Fictional narratives are unconstrained by reference to actual events: folktales, short stories, novels, and screenplays. Non-fictional narratives are constrained by reference to real events and people--journalism, documentary writing, and news reporting--where factual accuracy can in principle be verified. Personal narratives, such as first-person accounts on Reddit or oral histories collected in clinical settings recount lived experience and are non-fictional in intent, but unlike journalistic non-fiction their accuracy cannot be externally verified, which introduces concerns around self-presentation, reliability, and the relationship between narrative and identity that have been the primary object of sociolinguistic narrative analysis \citep{labov1967narrative}. \redtext{Elicited narratives, in contrast, are defined by their mode of production rather than their content: generated in response to a research stimulus under controlled conditions, they are methodologically distinct from naturally occurring discourse \citep{chelliah2010handbook}, and their prevalence in generation research reflects the practical difficulty of training and evaluating models on unconstrained narrative text.} These categories are not mutually exclusive: a crowdsourced story about a personal experience is simultaneously personal and elicited. The four-way distinction tracks which theoretical traditions are most applicable to a given dataset and because source type is correlated with task in existing research: generation research concentrates almost exclusively on elicited and fictional sources, while non-fictional and personal sources are the primary sites of understanding research.

\subsubsection{\redtext{Fiction}}


\redtext{Fictional narratives have been the primary object of narrative theories since Aristotle, making them a natural starting point for LLM-based research that draws on narratological frameworks. They are the most common source type in the surveyed literature (Table~\ref{tab:summary_counts}), spanning folktales, short stories, novels, plays, and film scripts. Their prevalence reflects both theoretical alignment--classical narratology was developed almost exclusively for fictional texts--and practical considerations: open-access repositories such as Project Gutenberg\footnote{Researchers should nonetheless exercise caution when using Project Gutenberg as a primary source: \citet{piper-2023-computational} warns that problems of sample selection and poor metadata can produce erroneous downstream claims, and recommends purpose-built research collections for cultural and historical analysis.} make large corpora freely available.}

Folktales enable cross-cultural analysis. For example, \cite{wu-etal-2023-cross} study 1,925 folktales from 27 cultures to examine moral values and biases, while \cite{hobson-etal-2024-story} use LLM as classifiers to investigate moral content across diverse genres. Specific collections are frequently employed for narrative modelling. The Aesop's Fables Dataset \citep{aesop_fables,ardiputra2021aesop} is used by \citet{chen-si-2024-reflections}, while FairytaleQA \citep{xu-etal-2022-fantastic} focuses on Question Answering from 278 children-friendly stories sourced from Project Gutenberg, used by \citet{xu2024fine}, \citet{subbiah2024reading}, and \citet{ahuja2025finding}. Additionally, \citet{gervas2024tagging} use synopses of 100 Russian folk tales for Propp character-function tagging with zero-shot and few-shot LLM prompting. The Propp Dataset \citep{finlayson2017propplearner} provides annotated Russian folktales used by \citep{wilmot2021memory} alongside the Shmoop Corpus for event segmentation and salience detection, pairing theory-derived narrative function labels with human-judged salience scores. Some short stories are sourced from Reddit, with user-generated fiction from responses to r/writingprompts \citep{xie-etal-2023-next}.


While sources such as short stories, including fairytales, fables, folktales, and contemporary fiction, are often chosen for their manageable length and accessibility \citep{lee_story_2023,chen-si-2024-reflections}, novels provide long-form texts with extensive character histories, making them ideal for character analysis and thematic studies \citep{piper-etal-2024-social,yang-anderson-2024-evaluating,hamilton-etal-2025-city}. For example, \cite{piper-wu-2025-evaluating} examine nineteenth-century novels to assess LLM performance in topic labelling, while \cite{bizzoni-feldkamp-2023-comparing} explore sentiment analysis in Hemingway’s \textit{The Old Man and the Sea}. Similarly, \cite{yang-anderson-2024-evaluating} benchmark computational features for representing character similarity across Jane Austen’s six major novels. For these traditional literary formats, researchers often draw from open-access repositories like Project Gutenberg\footnote{https://www.gutenberg.org/}. For instance, \citet{hatzel-biemann-2024-story} specifically use 30 detective novels from this source. It also provides the source texts for the Shmoop Corpus \citep{chaudhury2019shmoop}, which pairs 231 fictional works (145 novels, 62 plays, and 24 short stories) with summaries from the Shmoop website, used by \citet{wilmot2021memory} for salience detection. Furthermore, \citet{ran-etal-2025-bookworld} extract structured data from 16 English and Chinese novels, such as George R. R. Martin's \emph{A Song of Ice and Fire}, to simulate their fabula for understanding and generation. Dramatic works are similarly analysed; \bluetext{\citet{udrea2025speaks} evaluate scenes from 9 European plays, including Shakespeare's \emph{Hamlet} and Camil Petrescu's \emph{Danton}, accessed via a dataset constructed by \citet{fischer2019programmable}.} These approaches often align with distant reading traditions, which use statistical and visual techniques to reveal patterns in large texts \citep{moretti2005graphs,moretti2011network}. LLMs now address earlier challenges such as integrating close and distant reading and handling geospatial and temporal uncertainty \citep{janicke2015close}. Another valuable dataset is \citet{piper2025cr4}'s CR4-NarrEmote, a large-scale open-vocabulary narrative emotion dataset, comprising over 200,000 annotations from 3,738 crowdworkers across 43,000 passages of long-form fiction (and also non-fiction) spanning 150 years and twelve genres. Annotations align substantially with expert labels and are mapped to both dimensional (Valence-Arousal-Dominance) and categorical (NRC) emotion frameworks, providing a foundation for affective narrative analysis at scale.

Additionally, Wikipedia offers summarised narratives, including plot synopses of fictional works \citep{hatzel-biemann-2024-story}, film summaries \citep{tian-etal-2024-large-language} (and non-fiction accounts such as biographies, historical events, and geographical descriptions \citep{bartalesi2024using}). These are often used instead of the original because of their shorten lengths for economic reasons. Building on these summaries, the ``Tell me again! Multiple summaries for the same story" dataset \citep{hatzel2024tell} features story summaries from Wikipedia covering books, movies, TV, novels, plays, and biographies, and is used by \citet{hatzel-biemann-2024-story} to create narrative embeddings. Similarly sourced from Wikipedia is the Move Remakes Plots Dataset \citep{chaturvedi2018have}, used for finding story similarity \citep{hatzel-biemann-2024-story}.

Movie scripts provide dialogue-rich narratives for LLM research \citep{chun_aistorysimilarity_2024,shen2025adapting}, although their inclusion in LLM training corpora raises contamination concerns of the LLM having memorised the scripts \citep{tian-etal-2024-large-language}. \citet{chun_aistorysimilarity_2024} used 9 popular Hollywood film scripts as plain text with character names, dialogue, scene headings, and action annotations to detect similarity. NarraSum \citep{zhao2022narrasum} provides 122,000 document-summary pairs derived from plot synopses of movies and TV episodes spanning diverse genres; \citet{hobson-etal-2024-story} repurpose it as a source of fictional narratives for moral identification tasks, illustrating how large-scale summarisation datasets can be recontextualised for different purposes. The MovieGraph dataset \citep{papalampidi2019movie}, which annotates films with turning points drawn from the discourse-level structure of plot, is used by \citet{sterner-etal-2026-contrastive} for narrative salience modelling. \citet{shen2025adapting} draw concurrently on MovieGraph \citep{vicol2018moviegraphs} and MovieNet \citep{huang2020movienet} for mixed narrative understanding tasks in the film domain, separately targeting story-level actions and discourse-level structure. For transcriptions of audio formats, \citet{abdessamed2024identifying} perform narrative detection on the Spotify Podcast Dataset \citep{clifton2020100}, which contains fictional genres (e.g., audio drama). These datasets are particularly valuable for operationalising discourse-level theories of turning points and narrative arc within a domain rich in temporal and causal structure, and their use reflects the broader willingness of LLM-era research to extend narratological frameworks beyond print literature.

\subsubsection{\redtext{Non-Fiction}}

\redtext{Non-fictional narratives in the survey span a wider register range than any other source type, from literary non-fiction (e.g., biographies) to journalism, political communication, and social media. What unifies these varied forms is that the narrative makes claims about the real world, which sometimes shifts the research questions from those of literary analysis toward questions of framing, distribution, and reception. This orientation is reflected in the near-exclusive concentration of non-fiction papers in understanding tasks: researchers working with non-fictional sources are primarily concerned with how real events are narrativised and what social effects those narrativisations produce.}

\cite{piper-etal-2024-social} use crowd-sourced annotations and distant reading techniques on the CONLIT dataset \citep{piper2022conlit} to explore how social network structures are represented in contemporary non-fiction books (e.g., biographies) compared to fiction books (novels), and find that non-fictional narratives show higher rates of communication between characters, introduce significant characters later, feature more modular sub-networks which contribute to an overall network which is lower density and transitivity than fictional networks. Non-fictional narratives are also included in Wikipedia plot summaries of biographies from the ``Tell Me Again!" Dataset \citep{hatzel2024tell}, the diverse text types in NarraDetect \citep{piper2021narrative}, books from Project Gutenberg and non-fictional podcast transcripts \citep{clifton2020100}.


In the domain of non-fiction, news media serves as a critical resource. \citet{das-etal-2024-media} use the media frames corpus \citep{card-etal-2015-media} for articles on immigration and gun control from historical and contemporary US news; they propose a framework that extracts events and their relations to other events, and groups them into high-level situated narrative frames common across different news articles. \citet{heddaya-etal-2024-causal} use contemporary news data from the NOW Corpus \citet{davies2016now} and historical news data ProQuest. \citet{marino2024integrating} use 84,874 public news article URLs, titles, and descriptions from Meta's URL Shares Dataset for political discourse classification and clustering. The PolyNarrative (NEWS) dataset \citep{nikolaidis-etal-2025-polynarrative} is used by \citet{piskorski-etal-2025-semeval}. \cite{zhou-etal-2024-large} examine news from Mainland China, Hong Kong, Taiwan, the United States, and Canada to study shared beliefs and values, while \cite{hobson-etal-2024-story} use a global news dataset from Kaggle. News texts are also used for topic labelling tasks \citep{piper-wu-2025-evaluating}.

Beyond mainstream news, researchers have turned to datasets specifically designed to capture partisan, conspiratorial, and socially manipulative narratives. PartisanLens \citep{maggini2026partisanlens} is a multilingual dataset of hyperpartisan and conspiratorial immigration narratives drawn from European media, used for misinformation detection tasks operating at the discourse and situatedness levels. \bluetext{On social media platforms, the PHEME dataset \citep{zubiaga2017exploiting}, which aggregates rumours and their verifications from Twitter across multiple news events, and the COCO dataset \citep{langguth2023coco}, which annotates COVID-19 conspiracy theories circulated on the same platform, are used by \citet{liu2025raemollm} for cross-domain misinformation detection. These datasets extend non-fiction narrative analysis from traditional editorial media to digital discourse platforms, where narrative frameworks, such as emotional framing, character roles, and moral positioning, are used to identify manipulative or false information.} \redtext{Their presence in the survey reflects a broader shift in which narrative theories developed for literary and journalistic texts are increasingly applied to real-world social contexts.}

\subsubsection{\redtext{Personal and Elicited}}

Personal and elicited narratives both emerge from individuals rather than from institutional or literary traditions--the first organically, the second in response to a controlled research stimulus \citep{chelliah2010handbook, biber1993representativeness}. Personal narratives, drawn primarily from online platforms and clinical settings, bring with them the concerns of sociolinguistic and psychological narrative analysis: self-presentation, reliability, and the relationship between lived experience and its verbal reconstruction. The two categories frequently overlap: crowdsourced datasets such as HIPPOCORPUS \citep{sap2020recollection} contain both real and imagined personal stories, and many elicited corpora explicitly solicit first-person accounts of lived experience.

A significant portion of narrative datasets rely on user-generated personal content. Reddit provides personal narratives \citep{shen_heart-felt_2024,piper-bagga-2024-using,hobson-etal-2024-story} and discussions across diverse domains such as news, politics, fandom, technology, and relationships \citep{mire-etal-2024-empirical} are useful for extracting narrative features such as events, characters, plot, setting, problem, and conflict. AskReddit specifically serves as a source for the AskReddit Turning Points dataset \citep{ouyang2015modeling}, featuring user-generated non-fiction and personal texts annotated with \citet{labov1967narrative}'s orientation, action, and evaluation, and used for feature annotation by \citep{piper-bagga-2024-using, hobson-etal-2024-story}. A structurally distinctive Reddit source is the r/amitheasshole subreddit, used by \cite{brei-etal-2025-classifying} as part of their multi-source dataset for unreliable narrator classification; these first-person posts are notable in that narrators present events with an explicit aim of soliciting moral judgement, producing a form of motivated self-presentation whose unreliability is embedded in the communicative situation rather than only in textual features.

Other datasets capturing personal narratives include StorySeeker \citep{antoniak-etal-2024-people} which contains 33 categories of both stories and non-stories annotated with story and event spans and was expanded into STORYPERCEPTIONS \citep{mire-etal-2024-empirical} for feature annotation, as well as the TUNA (Texts with Unreliable Narrators) dataset \citep{brei-etal-2025-classifying}. Empathy in personal and elicited stories is explored through HIPPOCORPUS \citep{sap2020recollection}, which contains real and imagined stories from crowd workers alongside empathy annotations. Similar datasets include the Empathetic Stories Dataset \citep{shen-etal-2023-modeling,shen-etal-2024-empathicstories}, featuring 1,500 personal stories collected from online forums, video transcriptions, and experimental settings with empathy judgements from crowd workers, and the extended HeartfeltNarratives \citep{shen_heart-felt_2024}, which annotates narrative style features for their effect on empathetic responses. \citet{ma2026multi} sources narrative texts from Chinese Therapeutic Writing Samples, used for mental health prediction tasks and represents an instance where personal narrative is studied as a psychological and therapeutic practice rather than as a literary or social media artefact. Preprocessing across these datasets removes harmful or sensitive content, such as accounts of sexual assault or excessive profanity, relying on annotations provided by both human crowd workers and LLMs to reject harmful material \citep{shen_heart-felt_2024}.

A widely-used elicited narratives dataset is ROCStories \citep{mostafazadeh2016corpus}, consisting of crowdsourced, five-sentence stories designed to describe realistic events intended to capture common-sense causal and temporal relationships. It is accompanied by the Story Cloze Test, which is the task of differentiating between the right and wrong endings out of two potential endings. ROCStories has been used extensively, such as by \citet{hatzel-biemann-2024-story} for testing the accuracy of narrative embeddings, by \citet{mori2022plug} for emotion-aware story continuation guided by \citet{vonnegut_shapes_of_stories}'s affective arc shapes as generation targets, and by \citet{brei2024returning} for RENarGen, which fixes related opening and closing sentences and generates the middle of the story between them. It has also spawned numerous derivative datasets: \citet{rashkin2018modeling} annotated ROCStories with character mental states (e.g., motivations and emotions) as a new dataset called Story Commonsense which \cite{xie2022psychology} used for psychology-guided controllable story generation. \citet{gangal2022nareor} construct NAREORC, which are human rewritings of ROCStories five-sentence stories into non-linear narrative orders used to train and evaluate LLM reordering of narrative events. Additionally, the Choice of Plausible Event in Sequence (COPES) dataset \citep{wang-etal-2023-cola} functions as an annotated non-fiction subset of ROCStories, utilized by \citet{sun-etal-2024-event}. \bluetext{The TimeTravel dataset \citep{qin-etal-2019-counterfactual}, also derived from ROCStories, provides nearly 30,000 counterfactual story rewritings in which a single event is replaced by a counterfactual, with human-written alternative endings reconstructing a coherent narrative from the modified premise; it is repurposed by \citet{liu2026retell} as a training and evaluation resource for a reinforcement learning-based story generation system guided by narrative theory-derived reward scores.}


\subsubsection{\redtext{AI Generations}}

\redtext{AI-generated narratives form a distinct source category, we observe three use-cases: comparing LLM outputs against human-authored narratives; probing narrative biases within LLMs themselves; and supplying training data for model distillation.}

Used comparatively, AI-generated text is evaluated alongside human-authored narratives to examine specific narrative properties: \citet{beguvs2024experimental} compare human and machine-authored stories to examine how cultural schemas such as the Pygmalion myth are reproduced and varied across both; \citet{harel-canada-etal-2024-measuring} use AI-generated and human stories to develop and validate a framework for measuring psychological depth; and \citet{chhun-etal-2022-human} draw on both to construct a benchmark for evaluating story generation methods.

As probes of LLM bias, datasets intended hallucination evaluation have been used for narrative detection. \citet{sui_confabulation_2024} draw on FaithDial \citep{dziri-etal-2022-faithdial}, BEGIN \citep{dziri-etal-2022-evaluating}, and HaluEval \citep{li-etal-2023-halueval} to measure the narrativity of hallucinated versus factually grounded outputs; confabulated responses, they find, display significantly higher narrativity.

\bluetext{For distillation, AI-generated narratives serve as training data. \citet{xia2025storywriter} propose StoryWriter, a multi-agent framework comprising an outline agent, a planning agent, and a writing agent that coordinate to generate long-form stories of approximately 8,000 words. The system produces LongStory, a dataset of roughly 6,000 stories used for supervised fine-tuning of Llama3.1-8B and GLM4-9B--effectively distilling the narrative generation capacity of a multi-agent pipeline into smaller single models.}

\subsection{\redtext{Levels of Tasks}}
\label{sec:tasks_and_levels}

\begin{table}[!htbp]
    \footnotesize
    \caption{Narrative tasks and levels explored by NLP using LLMs}
    \label{tab:tasks_levels}
    \begin{tabularx}{\textwidth}{@{} >{\raggedright\arraybackslash}p{0.2\textwidth} @{\hspace{1em}} c @{\hspace{0.75em}} c @{\hspace{0.75em}} c @{\hspace{0.75em}} c @{\hspace{1.5em}} Y @{}}
        \toprule
        Tasks & \multicolumn{4}{c}{Level} & Explored By \\
        \cmidrule(l{1em}r{1.5em}){2-5}
        & \rotatebox{90}{Fabula} & \rotatebox{90}{Discourse} & \rotatebox{90}{Narration} & \rotatebox{90}{Situatedness} & \\
        \midrule
        Character Relationship identification
        & \checkmark & \checkmark &  &  & \citet{piper-etal-2024-social,hamilton-etal-2025-city,sancheti-rudinger-2025-tracking,udrea2025speaks}\\
        \bottomrule
        Fabula Simulation/Representation
        & \checkmark & \checkmark &  &  & \citet{ran-etal-2025-bookworld,yu-etal-2025-multi,yoo2024leveraging,chen2026storybox}\\
        \bottomrule
        Summarisation
        & \checkmark & \checkmark & & & \citet{chen-si-2024-reflections,subbiah2024reading,ahuja2025finding}\\
        \bottomrule
        Event Segmentation 
        & \checkmark & \checkmark &  &  & \citet{wilmot2021memory,bartalesi2024using,sun-etal-2024-event}\\
        \bottomrule
        Embedding 
        & \checkmark & \checkmark &  &  & \citet{hatzel-biemann-2024-story,sterner-etal-2026-contrastive}\\
        & \checkmark & &  &  & \citet{chun_aistorysimilarity_2024}\\
        \bottomrule
        Story Continuation (fiction)
        & \checkmark & \checkmark &  &  & \citet{mori2022plug,xie-etal-2023-next,ghaffari-hokamp-2025-narrative,bissell-etal-2025-theoretical,liu2026retell}\\
        \bottomrule
        Controlled Story Generation (fiction)
        & \checkmark & \checkmark &  &  & \citet{gangal2022nareor,xie2022psychology,mirowski2023co,brei2024returning,ermolaeva-etal-2024-tame,lima2024pattern,song2024conflict,venkatraman2025collabstory,wang2025generating,xia2025storywriter}\\ 
        \bottomrule
        Narrative Detection or Extraction 
        & \checkmark & \checkmark & & & \citet{abdessamed2024identifying,antoniak-etal-2024-people,sui_confabulation_2024} \\
        & \checkmark & \checkmark & & \checkmark & \citet{piper_narradetect_2025,mire-etal-2024-empirical} \\
        & \checkmark & \checkmark & & \checkmark & \citet{marino2024integrating,nikolaidis-etal-2025-polynarrative,piskorski-etal-2025-semeval,schneider2026llm,maggini2026partisanlens}. \\
        & \checkmark & \checkmark & & \checkmark & \citet{das-etal-2024-media,heddaya-etal-2024-causal}. \\
        \bottomrule
        Moral Identification 
        & \checkmark & \checkmark & & \checkmark & \citet{hobson-etal-2024-story,zhou-etal-2024-large} \\
        \bottomrule
        Character Role identification
        &\checkmark &\checkmark &  &  & \citet{gervas2024tagging,yang-anderson-2024-evaluating} \\
        & \checkmark & \checkmark &  & \checkmark & \citet{piskorski-etal-2025-semeval}\\
        \bottomrule
        Discourse Feature identification
        & & \checkmark &  &  & \citet{piper-bagga-2024-using}\\
        & \checkmark & \checkmark & & \checkmark & \citet{bissell-etal-2025-theoretical}\\
        & \checkmark & \checkmark & \checkmark & \checkmark & \citet{shen_heart-felt_2024,brei-etal-2025-classifying}\\
        \bottomrule
        Story Quality Criteria
        & \checkmark & \checkmark &  & \checkmark & \citet{chhun-etal-2022-human,chakrabarty2024art,yang-jin-2025-matters}\\
        && \checkmark &&\checkmark & \citet{harel-canada-etal-2024-measuring}\\
        \bottomrule
        
        Misinformation Detection
        & \checkmark & \checkmark & & \checkmark & \citet{liu2025raemollm,maggini2026partisanlens}\\
        \bottomrule
        Controlled Story Generation (non-fiction) 
        & \checkmark & \checkmark &  & \checkmark & \citet{hamilton2022covid,vykopal2024disinformation}\\
        \bottomrule
    \end{tabularx}
\end{table}

As discussed in Section \ref{sec_narrative_landscape}, narrative processing spans multiple levels of abstraction. At the fabula level, it involves tasks such as event segmentation, character classification, interaction classification, and event construction. The discourse level focuses on the extraction and annotation of textual features. The narration level encompasses tasks such as understanding rhetorical intention and reliability classification. The situatedness level addresses tasks concerned with how narratives are produced, received, and circulated in specific socio-cultural contexts, including narrative detection in political communication and misinformation identification. Several tasks cut across these levels, including narrative embedding, generation, narrativity measurement, psychological depth evaluation, cultural bias assessment, narrative reconstruction, and similarity detection.

Since the target of NLP is always the text itself, all tasks are ultimately grounded in discourse. Even tasks primarily concerned with the fabula such as event segmentation and character relationship identification still require the text as their starting point, since chronological events and character interactions must be extracted from textual presentations of the story world.

Within generation specifically, a recurring architectural pattern applies the fabula/discourse distinction directly: a two-stage pipeline in which a fabula-level plan is generated first (events, characters, goals, causal structure) before discourse-level prose is produced from it. This design--visible in systems as varied as \cite{ran-etal-2025-bookworld,yu-etal-2025-multi,chen2026storybox,yoo2024leveraging}--suggests that the classical narratological separation of story world from its textual presentation is used as a methodological principle for LLM-based story generation. It also echoes what earlier surveys \citep{alhussain2021automatic,wang2023open} termed planning-based approaches, \redtext{showing that the planning/generation distinction persists in LLM-based story generation methods.}

Almost every generation task in the survey--controlled story generation, story continuation--operates exclusively at the fabula and discourse levels. Apart from two examples--\citet{hamilton2022covid} on pandemic new framing and \citet{vykopal2024disinformation} on disinformation generation, no generation task reaches the narration or situatedness levels. For exmaple, \citet{vykopal2024disinformation} investigate the generation side of situated narrative by examining the disinformation capabilities of LLMs, finding that models can generate contextually plausible false narratives at scale, explicitly addressing the real-world reception and effect of produced narratives. 

This reflects a fundamental constraint of current LLM-based generation: systems model what happens in a story (fabula) and how it is arranged and expressed (discourse), but do not yet produce narratives that explicitly position a narrator relative to the diegesis, or that are designed for a specific situated context of production and reception. By contrast, some understanding tasks span all four levels. The implication is that while LLMs can be prompted or fine-tuned to recognise higher-order narrative properties--narrator reliability \citep{brei-etal-2025-classifying}, moral framing \citep{hobson-etal-2024-story}, the contextual circulation of narratives in political media \citep{nikolaidis-etal-2025-polynarrative}--the situatedness of narratives has not been the focus of generative modelling.

Tasks that include understanding narrative situatedness are more commonly applied to non-fictional datasets--misinformation detection, moral identification in news media, narrative extraction from political discourse--and are less prominent tasks applied to literary fiction. This is consistent with the nature of situatedness as a concern with how narratives are produced, distributed and received in specific socio-cultural contexts \citep{herman1997scripts}, which is more commonly addressed in research treating narrative as social practice than in research treating it as textual artefact.

Character-focused tasks--character relationship identification and character role identification--cluster predominantly at the fabula level, consistent with the narratological understanding of characters as belonging to the story world rather than its textual presentation. The exception, visible in \citet{piskorski-etal-2025-semeval}'s work on the PolyNarrative dataset, extends character role analysis to discourse and situatedness in news contexts, where the narrative function of social actors is shaped not only by story-world logic but by ideological and rhetorical positioning. Misinformation and disinformation tasks, by contrast, uniquely span from discourse upward to narration and situatedness. This reflects the nature of disinformation as an inherently situated narrative act whose effectiveness depends not only on what is said and how, but on the relationship between a narrator (real or implied), an intended narratee, and the broader context of reception--precisely the concerns of narration and situatedness as narrative levels.

Story quality evaluation tasks span fabula, discourse, and situatedness. The multi-dimensionality of narrative quality evaluation--which must attend simultaneously to what happens in a story, how it is told, and its situated context--partially explains why standardised benchmarks for story quality remain elusive, since it is difficult for a single benchmark to capture all these dimensions.

\section{Narrative Theories in the modern NLP pipeline}\label{sec_NLP_pipeline}

This section examines how narrative theories enter the three stages of the NLP pipeline: labelling, modelling, and evaluation. Labelling refers to the human or automated annotation of datasets; modelling involves training or prompting LLMs for a specific task; and evaluation is the qualitative or quantitative assessment of model outputs. While existing surveys tend to treat narrative theory as a source of motivation or context for NLP work, our survey shows that narrative theories actively shape what gets labelled, how models are structured, and what counts as a good output. Through examining the surveyed papers, we find that theories enter the labelling stage as annotation schemes that determine label format and granularity; they enter the modelling stage as prompt instruction content, structured input conditions, pipeline architectures, and training objectives; and they enter the evaluation stage as sources of criteria against which outputs are judged. LLMs appear at every stage, sometimes as the primary system being evaluated, sometimes as annotators or judges, and sometimes at multiple stages simultaneously. Further specifics of LLM methods are examined in Section~\ref{sec_LLM_techniques}.

\subsection{Labelling}\label{sec7_1_1}

\redtext{The labelling stage converts narrative texts into annotated datasets by applying theoretical frameworks as label schemes, where format of those labels shapes what a downstream model can learn. Across the surveyed work, labels take three broad forms: span-level annotations that segment texts into functional units, categorical labels that classify elements or entities, and ordinal ratings that quantify abstract narrative properties.}

Span-level annotations divide texts into segments. \cite{troiano_clause-atlas_2024} draw from socio-linguistic narrative theory \citep{labov1972language,berman1997narrative} to annotate each clause in 19th and 20th century novels as event, subjective experience, or contextual information, identifying experiencers for subjective clauses. This clause-level segmentation produces a structured representation of narrative sequence that downstream models can use for classification and structural pattern detection.

Categorical labels classify narrative elements into discrete types. \cite{piper2022toward} label texts from diverse genres--short stories, novels, Reddit posts, scientific abstracts, and legal contracts--using features such as temporality, eventfulness, and agenthood to assign narrativity categories. \cite{yang-anderson-2024-evaluating} group Jane Austen characters into five social categories (gender, age, rank, wealth, marital status) and seven narrative roles (e.g., heroine, hero, deceiver). \cite{piper-etal-2024-social} classify character interactions in the CONLIT dataset into five types: communicating, thinking, observing, touching, and associating. \citet{brei2026casper} apply \citet{hochman1985character}'s eight-dimension character taxonomy--stylisation, coherence, wholeness, literalness, complexity, transparency, dynamism, and closure--to classify protagonists in human-written and LLM-generated stories. While many frameworks rely on these predefined categories, recent approaches are pushing toward open-ended annotations. For instance, Taxonomy-Free Character Role Labeling (TF-CRL) uses LLMs to assign compositional, open-vocabulary labels (e.g., `Resilient Leader' or `Scapegoated Visionary') to actors in news stories, capturing more nuanced functional roles than fixed taxonomies allow \citep{hobson2025evaluating}

Ordinal ratings quantify abstract narrative properties on a defined scale, producing scalar scores that allow narrative features to be compared systematically across texts and correlated with reader responses. Drawing on cognitive narratology, \cite{bissell-etal-2025-theoretical} ask trained annotators to rate story endings on six surprise-related criteria using a 5-point Likert scale \citep{ortony1987surprisingness,brewer1980event,celle2017expressing}. \cite{shen_heart-felt_2024} develop a taxonomy based on theories of narrative empathy \citep{keen2006theory,van2017evoking} to guide expert annotations of personal stories. \cite{chhun-etal-2022-human} use criteria derived from psycholinguistics \citep{mccabe1984makes}, business management \citep{dickman2003four}, and interactive digital storytelling \citep{bae2021preliminary} to rate human and AI-generated short stories on relevance, coherence, empathy, surprise, engagement, and complexity.

Labels can be applied by expert annotators, crowd workers, or LLMs. LLMs are increasingly used to apply theory-derived labels at scale, with validity assessed by comparison with human annotations \citep{chun_aistorysimilarity_2024,piper_narradetect_2025,michelmann_large_2025}. \cite{tian-etal-2024-large-language} provide both expert and automated annotations for story arcs and turning points. There are claims that output variability from different instruction wordings is comparable to variation among human annotators, and that labels produced by different prompts can approximate the range of human annotation behaviour \citep{troiano_clause-atlas_2024,harel-canada-etal-2024-measuring,chhun2024language,zhou-etal-2024-large}. Reported correlations between LLM and human annotations are typically low to moderate \citep{chhun-etal-2022-human,sancheti-rudinger-2025-tracking,deering-penn-2025-analysis}, a finding that recurs across tasks and models and suggests that LLM-based labelling at scale is not yet always a reliable substitute for expert annotation for narrative tasks.

\subsection{Modeling}
\label{sec_modeling}

At the modeling stage, pre-trained language models offer flexibility across a range of tasks, which can broadly be divided into understanding and generation. In this survey, we classify narrative generation tasks as those requiring the output to be a narrative, while understanding tasks output annotations, summaries or classifications. Some tasks combine both, such as iteratively evaluating and improving narrative generation \citep{chen-si-2024-reflections}.

\subsubsection{\redtext{Modelling Understanding}}

In understanding tasks, the input is an existing narrative text and the output is a representation of some narrative property extracted from it, where the form of that output reflects the theoretical framework being applied. Across the surveyed understanding papers, the range of properties extracted spans from surface-level features such as event sequences and character types to increasingly abstract constructs including narrator reliability, moral schemas, and cultural bias.

Frameworks that treat narrative properties as continuous dimensions produce scalar scores: for example, narrativity measured as a continuous probability score across a corpus of hallucinated versus factual outputs \citep{sui_confabulation_2024}, or story quality dimensions such as coherence, empathy, and surprise rated on Likert scales \citep{chhun-etal-2022-human}. These scores make theoretically defined properties comparable across texts at corpus scale, enabling the kind of distributional analysis that close reading cannot support.

Frameworks that define discrete narrative types produce categorical labels: clause-level narrative function categories--eventive, subjective, or contextual information--derived from sociolinguistic narrative theory \citep{troiano_clause-atlas_2024}; character interaction types such as communicating, observing, or touching, derived from network theory \citep{piper-etal-2024-social}; cross-genre moral schemas \citep{hobson-etal-2024-story}; narrator unreliability categories \citep{brei-etal-2025-classifying}; and character category dimensions \citep{brei2026casper}. 


Frameworks that model narrative as a system of relations produce graph-based structures: social networks extracted from multilingual fiction and non-fiction \citep{hamilton-etal-2025-city}, event-centric narrative graphs from news \citep{das-etal-2024-media}, and causal micro-narrative graphs from news corpora \citep{heddaya-etal-2024-causal}, which cluster almost exclusively at the fabula level, consistent with the narratological treatment of events and characters as belonging to the story world rather than its textual presentation.

Finally, some frameworks produce vector embeddings that encode narrative structure: unlike generic semantic embeddings, which treat all surface variation as signal, narrative-specific embeddings abstract away discourse-level variation to preserve fabula structure--a distinction that maps directly onto the classical fabula/discourse binary. The theoretical framework thus determines not just what is measured but what counts as noise: \citet{hatzel-biemann-2024-story} train embeddings that place reformulations of the same story closer together than texts with different plots, while \citet{sterner-etal-2026-contrastive} produce contrastive embeddings trained on story pairs that share fabula but differ in discourse representations.

\subsubsection{\redtext{Modelling Generation}}
\redtext{In generation tasks, the system transforms the input into an output narrative text. Across the surveyed generation papers, three patterns emerge by which theory enters as input into different systems.}

The most common of these embeds narrative theory as prompt content, writing theory-derived categories or structural templates directly into the instruction prompt. \citet{halperin2024artificial} integrate Propp's story functions and the Aristotelian three-act structure into planning prompts. \citet{mirowski2023co} decompose scriptwriting into linked hierarchical prompts whose stages echo Aristotelian dramatic categories. \citet{ermolaeva-etal-2024-tame} supply Proppian narrative functions as prompt-controlled stage transitions in a fairy-tale generation framework. \citet{tian-etal-2024-large-language} write a target story arc type from a \citet{vonnegut_shapes_of_stories}-derived schema directly into the generation prompt, and separately embed turning point definitions into a planning outline that the model then expands into a full story.

Other systems instead treat theory as a structured input condition, where theory defines variables that are computed separately and given to the model as explicit inputs rather than as instruction text. \citet{xie2022psychology} give a fine-tuned BART model a per-character psychological state chain--derived from \citet{maslow1943theory}'s hierarchy of needs and \citet{plutchik1980general}'s wheel of emotions--as a conditioning input that the generated story must satisfy at each event. \citet{song2024conflict} operationalise \citet{freytag1894technik}'s notion of conflict as a computed obstacle that reduces the probability of the protagonist achieving their goal, passing it as a structured \texttt{(Context, Goal, Obstacle)} tuple to a fine-tuned GPT-2 for story completion. \citet{gangal2022nareor} gives BART, T5, and GPT-2 an explicit target narrative order as input, training models to rewrite a five-sentence story into a specified non-linear discourse arrangement while preserving its fabula. \citet{brei2024returning}'s RENarGen first produces an ending sentence that relates back to the opening, then fills in the middle conditioned on both endpoints--producing stories rated as having stronger closure or finality \citep{carroll2007narrative} than left-to-right autoregressive baselines.

The theory can sometimes shape the pipeline architecture, determining the sequence of generation stages rather than the content of any single prompt. \citet{mirowski2023co}'s Dramatron bases its pipeline design on Aristotle's identification of plot, theme, character, and dialogue as distinct dramatic elements, sequencing generation from logline through characters, plot beats, location descriptions, and dialogue. The theory is not mentioned in any individual prompt; it is the rationale for the architecture itself. Multi-agent systems represent the main structural innovation in this pattern and most directly instantiate it through the fabula/discourse distinction, which maps onto distinct agent roles. \citet{ran-etal-2025-bookworld}'s BOOKWORLD uses character agents to simulate fabula-level events, whose output feeds a separate agent that produces prose. \citet{yu-etal-2025-multi} implement a Director Agent that coordinates character agents to generate fabula content, which a Rewrite Agent then transforms into finished story text. \citet{chen2026storybox} follow the same design. In each case, the fabula/discourse distinction maps directly onto distinct input/output boundaries between agents.

\subsubsection{\redtext{Modelling Both}}

\redtext{Some tasks combine understanding and generation within the same pipeline, where the output of one process serves as input to the other. The form of this coupling varies: in some systems, understanding produces an intermediate representation that generation then operates over; in others, the two processes are interleaved in an iterative loop; in reinforcement learning with AI feedback, understanding produces a score that feeds into a training objective for the generator.}

\citet{chen-si-2024-reflections} address the task of story reconstruction--generating a version of a source story that preserves its underlying narrative structure--using a two-agent system applied to short story summaries. A Reflection agent takes the source text as input and extracts narrative properties from it: characters' psychological states, temporal structure, and literary techniques. Those extracted properties are passed as structured input to a Resonance agent that generates a reconstruction and compares it against the source, with the comparison score feeding back to refine the Reflection agent's extraction prompts in the next iteration. 

\citet{ghaffari-hokamp-2025-narrative}'s Narrative Studio takes an existing story as input and constructs a causal event graph of its actors, environment, and state transformations, grounded in \citet{todorov19712}'s theory of narrative equilibrium. That graph is not an end product but an input to generation: Monte Carlo Tree Search navigates the graph's possible state transitions and invokes LLM inference at each node to generate and evaluate story continuations, expanding the branches that best satisfy user-defined narrative criteria. 

\citet{liu2026retell} take a source story and counterfactual premise as input; a generated retelling is passed to an LLM-as-judge that evaluates it against \citet{todorov19712}'s narrative equilibrium principles and returns a reward score; that score drives a GRPO reinforcement learning loop with LoRA adapters, combining understanding and generation through the training process rather than through inference.

\subsection{Evaluation}\label{sec7_1_3}

\redtext{Evaluation is the stage at which model outputs are assessed against some standard of narrative quality, and the choice of evaluation method reflects a theoretical commitment about how narrative quality is determined. Understanding tasks produce discrete outputs--categorical labels, ordinal scores, or structured representations--that can be directly compared against human annotations using standard classification or agreement metrics (Cohen's kappa, Krippendorff's alpha, F1, Spearman's correlation).}

While standard metrics can evaluate individual understanding tasks in isolation, the question of which narrative understanding capabilities LLMs actually possess across levels and genres remains poorly benchmarked. \citet{hamilton-etal-2026-narrabench} survey 78 existing benchmarks and find that only 27\% of narrative tasks are well-captured, with narrative events, style, perspective, and revelation nearly absent from current evaluations. Established benchmarks such as FairytaleQA \citep{xu-etal-2022-fantastic}--question answering across 278 children's stories sourced from Project Gutenberg--are confined to fabula-level comprehension, leaving discourse and narration-level phenomena largely unevaluated. Others such as \citet{ahuja2025finding}'s FlawedFictions benchmark tests plot hole detection as a measure of narrative consistency, finding that performance degrades sharply with story length and that LLM-generated summaries and stories introduce over 50\% and 100\% more plot holes respectively than human-written originals. 

Generation evaluation is complicated. Reference-based methods compare model outputs against gold standards using metrics such as BLEU, and BERTScore \citep{lee_story_2023,chen-si-2024-reflections}, but these correlate poorly with human judgements and tend to underestimate LLM performance \citep{qin-etal-2019-counterfactual}. Narrative generation is particularly resistant to reference-based evaluation because there is no single correct way to tell a story: a generated narrative may be coherent and engaging while differing entirely from any ground truth or gold standard. Lexical metrics for narrative tasks are a poor fit, and standardised benchmarks remain elusive. We believe this is not only a technical limitation but a consequence of narrative phenomena being distributed across multiple levels of abstraction (fabula, discourse, narration, situatedness) that no single reference text can capture.

Different theories are often combined into evaluation frameworks. \citet{chakrabarty2024art}, drawing on \citet{rodriguez2008problem}, decompose narrative into elements--plot, discourse-time, character, setting, and narration--as evaluation criteria, echoing narratological frameworks associated with \citet{chatman1980story}. \citet{yang-jin-2025-matters} draw on a wider range of inspirations, grounding their evaluation dimensions for book-length story assessment in theories of plot, structure, character, rhetoric, and empathy \citep{genette1980narrative,campbell2008hero,phelan1996narrative,mckee1997substance,booth1961rhetoric,keen2006theory}. Qualitative methods include assessing narrative elements such as plot, structure, character motivation, and setting according to rubrics adapted from creative writing assessment \citep{beguvs2024experimental,halperin2024artificial}. The Torrance Tests of Creative Writing (TTCW), proposed by \citet{chakrabarty2024art} and adapted from the Torrance Tests of Creative Thinking, evaluate creativity across 14 binary tests organised into dimensions of Fluency, Flexibility, Originality, and Elaboration; LLM-generated stories pass these tests 3--10 times less often than stories by professional authors. These cases illustrate that in the evaluation of LLM generated stories, narrative theory functions less as a single unified framework and more as a repertoire of concepts from which evaluation criteria are drawn, which mirrors the lack of consensus in narrative studies itself about what constitutes a good story.

Statistical analyses measure specific features of narrative texts, such as word category frequencies, demographic representations, or affective properties. \citet{beguvs2024experimental} apply this approach to compare 250 human-authored and 80 GPT-generated stories written to identical prompts: by counting the gender and racial identities of characters in each story and applying inferential statistics to compare distributions, they find GPT-4 stories represent gender roles significantly more progressively than human-authored texts. \citet{tian-etal-2024-large-language} use the NRC-VAD lexicon \citep{mohammad2018obtaining} to compute token-level arousal and valence scores aggregated sentence by sentence across LLM- and human-written stories, finding that LLM-generated stories cluster at high positive valence with low arousal throughout--homogeneously positive and structurally flat relative to human writing.

Human evaluation of generated narratives takes several distinct forms depending on what is being measured. Pairwise comparison asks evaluators to judge which of two stories is better on a given criterion; \citet{ran-etal-2025-bookworld} use this approach to evaluate BookWorld-generated stories, obtaining a 75.36\% win rate over baseline methods. Likert-scale rating asks evaluators to score individual stories on specified dimensions; \citet{chhun-etal-2022-human} have crowd workers rate 1,056 generated stories on six criteria (Relevance, Coherence, Empathy, Surprise, Engagement, Complexity) on 5-point scales. Ranked-choice voting presents evaluators with multiple options to rank rather than rate; \citet{bissell-etal-2025-theoretical} use this approach to have readers rank story endings across 120 narratives for narrative surprise criteria. \citet{chakrabarty2024art} recruit 10 creative writers to apply 14 binary TTCW tests to 48 stories. Expert annotation tasks trained annotators with domain knowledge to identify specific narrative features; \citet{tian-etal-2024-large-language} recruit 16 annotators to classify story arcs and identify turning point sentences, then separately rate generated stories for suspense, emotion provocation, and overall preference. \citet{subbiah2024reading} work directly with story authors as evaluators of LLM-generated summaries of their own unpublished work, finding faithfulness mistakes in over 50\% of summaries.

\subsubsection{LLM as Evaluator} 

\redtext{LLMs themselves are increasingly used as proxies for human evaluators to assess LLM outputs. LLM-as-evaluator is appealing for narrative tasks because it does not require a gold standard or ground truth comparison text, and its judgement of narrative quality is generalisable to different criteria and contexts. Studies generally find LLMs tend to outperform lexical metrics such as BLEU and BERTScore, and reach low to moderate alignment between LLM-evaluators and humans, with reliability varying by criteria and/or task type. Some studies \citep{yu-etal-2025-multi} already rely on LLM-as-evaluator exclusively, reflecting growing practical acceptance despite its limitations.} 


Some examples in the surveyed literature demonstrate that LLM-evaluator alignment can be improved for narrative generation evaluation via pipelines, careful model selection, and fine-tuning. \citet{sun-etal-2024-event} show that prompting an LLM to first generate a causal event graph of a story before producing an overall quality score yields 3.6--16.6\% relative improvement in correlation with human ratings. \citet{liu2026retell} test three candidate LLM judges--Selene-1-mini-8B, M-Prometheus-14B, and Gemini-3-Flash--against human annotators on 200 stories using Todorov-derived criteria, finding fair to moderate agreement, and select Gemini-3-Flash as the final evaluator on the basis of its best alignment with human ratings. \citet{yang-jin-2025-matters} train a specialised 8B model (NovelCritique) on reader reviews to evaluate book-length stories, finding it outperforms GPT-4o in alignment with human ratings--suggesting that domain-specific training data may be more effective than model scale for closing the gap between LLM and human narrative judgement. However, \citet{deering-penn-2025-analysis} find that fine-tuning a model on human story ratings offers no advantage over simply prompting an LLM to rate stories directly: all three scoring methods they compare (log-likelihood, zero-shot prompting, and fine-tuning) produce similarly average results when evaluated by human raters.

The evidence on LLM-as-evaluator reliability is mixed and dependent on task/criteria. \citet{chhun2024language} conduct a systematic analysis, rating 1,056 stories across six criteria (Relevance, Coherence, Empathy, Surprise, Engagement, and Complexity) using four prompt variants. LLM ratings show high correlation with human judgements when ranking story generation systems against each other, though this reliability does not extend to individual story evaluation, where correlations are moderate to weak. They find that prompt complexity does not improve human alignment, and LLMs produce consistent ratings across prompt variants but fail to explain them with substantiated reasoning. \citet{harel-canada-etal-2024-measuring} find strong human inter-rater agreement on their five-dimension Psychological Depth Scale (0.72 Krippendorff's alpha), and LLM-human correlation averages only 0.51 and varies considerably across dimensions, peaking at 0.68 for empathy. \citet{chakrabarty2024art} use the Torrance Tests of Creative Writing (TTCW)--which evaluates fictional stories across Fluency, Flexibility, Originality, and Elaboration--to examine whether ChatGPT, GPT-4, and Claude 1.3 align with expert human assessments of story quality. All three models correlate with expert judgements close to zero across all four dimensions, but this finding requires context: human expert agreement on the same tests is itself only moderate (Fleiss Kappa 0.41, ranging from 0.27 to 0.66 across individual tests). Evaluating creative writing is inherently subjective, and the degree to which any evaluator (human or LLM) can be considered reliable depends on a consensus that does not always exist.

\section{LLM Techniques for Narrative Tasks}\label{sec_LLM_techniques}


\redtext{This section explores the LLM specific approaches for narrative tasks: pre-training, post-training, in-context learning and compound systems and examine how they are used in various methodologies. We find that prompting-only in-context approaches currently predominate research methodologies, potentially due to their simplicity or, alternatively, indicative of the adaptability of modern pre-trained language models to various tasks.} While encoder-only models such as BERT are still used by researchers for story understanding tasks \citep{baumann-etal-2024-bert,sui-etal-2023-mrs}, decoder LLMs can be used in a variety of tasks such as generation, feature analysis and classification. Decoder LLMs have significant advantages: 1) They can connect the low level linguistic features captured by traditional methods in NLP and high level theoretical constructs. 2) They can detect a variety of narrative features at large-scale where without the need of abundant training data. 3) They introduce accessibility of use for interdisciplinary adoption \citep{piper_narradetect_2025}. 

State-of-the-art propriety pre-trained language models such as GPT, Gemini, and Claude have been chosen to carry various understanding and generation tasks \citep{shen_heart-felt_2024,troiano_clause-atlas_2024}. \cite{piper-bagga-2024-using} state their concern over replicability when using propriety models like GPT-4, and consider \redtext{open-weight} models to be valuable for replicability and benchmarking. For open-weight models, researchers tend to use low-parameter models (like 8B) on limited hardware \citep{tian-etal-2024-large-language,piper-etal-2024-social,hatzel-biemann-2024-story,piper-wu-2025-evaluating} with quantisation (which reduces the precision of the model's numerical values such as by converting floats to integers) further reducing memory usage \citep{piper-bagga-2024-using}. 

\subsection{Pre-training}

Overall, researchers almost never conduct unsupervised learning and pre-train LLMs on large volumes of narrative texts themselves. While there is yet to be a study which investigates the exact presence of narrative text in LLM pre-training data, we can induce its presence in confirmed pre-training sources such as through books \citep{zhu2015aligning}, Wikipedia \citep{hadad2026wikipedia}, news and blogs \citep{myntti2025register}.


Parametric memory--knowledge encoded in model weights during pre-training rather than supplied at inference--functions as either a resource or a contamination risk depending on the research goal. \citet{chun_aistorysimilarity_2024} exploit it deliberately in the context of story similarity: rather than providing film scripts as input, they query the model directly to extract narrative elements from its parametric memory for comparison against human judgements, positioning parametric memory as an alternative to context-based evaluation. \citet{tian-etal-2024-large-language}, by contrast, treat pre-training exposure as a confound to be controlled: their evaluation dataset draws exclusively from 2020s Wikipedia film synopses, with titles rephrased and synopses filtered by length to reduce contamination risk. Both cases are complicated by a lack of training data transparency: for proprietary and most open-weight models, whether a given text appeared in pre-training is unknowable, making parametric influence difficult to verify or control. Fully open models such as OLMo \citep{groeneveld-etal-2024-olmo}, on the other hand, allow researchers to audit pre-training corpora directly.

Continuous pre-training for domain adaptation has been found to be useful in finance \citep{xie-etal-2024-efficient}, medicine and law \citep{guo-etal-2025-efficient-domain}. The only example of pre-training as a narrative processing methodology we found is \cite{meaney-etal-2024-testing}'s (non-narrative theory informed and non-LLM) experiment with domain adaptation which applies continuous pre-training with Irish and Gaelic language folktales dataset to BERT based encoder-only models for the tasks of classifying the narrator's gender and the type of the folktale. 


We hypothesise that LLMs inherit narrative biases from pre-training. \citet{sui_confabulation_2024} provide indirect but suggestive evidence: hallucinated outputs display significantly higher narrativity than factually accurate ones, implying that when LLMs generate without grounding they default to narrative structure--consistent with the hypothesis that narrative structure is encoded in model weights. \bluetext{Further evidence comes from \citet{zhou2026amory}, who find that organising agent memory as episodic narratives outperforms embedding-based RAG on long-term reasoning tasks, suggesting LLMs retrieve narrative representations more effectively than alternatives. Despite this, how the quantity and type of narrative content in pre-training shapes LLM behaviour on narrative tasks remains largely uninvestigated, and we identify this as a gap in the literature.}

\subsection{\redtext{Post-Training}}\label{finetune}

\redtext{Post-training is common and effective for story understanding but rare and constrained for story generation. Methods across the surveyed literature include, supervised fine-tuning (SFT), model distillation, contrastive learning, and reinforcement learning--but their application is heavily skewed toward understanding tasks because it is easier to define what a correct annotation is than what a good story is.}


Supervised fine-tuning has been shown to be effective for various narrative understanding tasks across the surveyed papers: story quality evaluation trained on 176k novel reviews \citep{yang-jin-2025-matters}, narrative-specific embeddings trained on same-plot story pairs \citep{hatzel-biemann-2024-story}, movie-domain narrative understanding trained on \citet{vicol2018moviegraphs}'s MovieGraph and \citet{huang2020movienet}'s MovieNet \citep{shen2025adapting}, and character interaction classification on crowd-annotated passages \citep{piper-etal-2024-social}. Fine-tuned smaller models can outperform larger untuned ones: \citet{sui_confabulation_2024}'s fine-tuned ELECTRA-large outperforms GPT-4 zero-shot on narrative detection, and \citet{yang-jin-2025-matters}'s NovelCritique--an 8B model fine-tuned on reader reviews--outperforms GPT-4o on alignment with human story evaluation ratings. \citet{brei-etal-2025-classifying} apply curriculum learning--a training strategy that introduces tasks in progressively increasing order of difficulty--to unreliable narrator classification, one of the few tasks in this survey targeting the narration level. Training consistently outperforms zero-shot, though performance degrades on the inter-textual task, which requires inference about narrator intent beyond surface textual cues.

Model distillation sees large proprietary models generate annotations that are then used to train smaller open-weight models. \citet{piper-bagga-2024-using} use GPT-4 outputs to fine-tune Llama3 (8B), Mistral (7B), and Mixtral (56B) with LoRA on narrative discourse features derived from \citet{genette1980narrative} and \citet{herman2009basic}. \citet{xia2025storywriter} apply the distillation to generation: a multi-agent system produces LongStory, a dataset of approximately 6,000 stories averaging 8,000 words each, which is then used to fine-tune Llama3.1-8B and GLM4-9B--distilling the narrative generation capacity of the multi-agent system into smaller models.

\citet{sterner-etal-2026-contrastive} train contrastive story embeddings on narrative twins--pairs sharing the same plot but differing in surface form--alongside LLM-generated distractors with similar surface features but different plots, operationalising the Labovian notion that salient events advance rather than merely describe a narrative. Evaluated across four operations (deletion, shifting, disruption, and summarisation) on ROCStories and Wikipedia plot summaries, the embeddings outperform masked-language-model baselines, with summarisation the most reliable for inferring salience.

Post-training as a primary method\footnote{Some methods include post-training as a sub-component within a compound pipeline--including \citet{xie2022psychology}, \citet{brei2024returning}, \citet{song2024conflict}, and \citet{xia2025storywriter}. These methods do not primarily rely on training to improve generation quality.} for narrative generation is rare. \citet{xie-etal-2023-next} established an early finding that GPT-3 with few-shot prompting outperformed fine-tuned models including BART, KGGPT2, and HINT. As subsequent attempt at open-ended supervised fine-tuning for generation, \citet{seredina2024report} trained model on Project Gutenberg novels and produced outputs that were unable to make smooth transitions between chapters. Generation lacks the bounded annotation schema that makes fine-tuning tractable for understanding tasks, and the absence of agreed-upon quality metrics makes reward modelling difficult outside narrowly defined theoretical constraints. We found only three papers apply standalone post-training for generation. \citet{gangal2022nareor}'s NAREOR fine-tunes BART, T5, and GPT-2 on NAREORC--over 1,000 human rewritings of ROCStories into non-linear narrative orders--where the model takes a story and a target narrative order as input and must produce a rewrite that preserves the fabula while presenting it in the specified sujet arrangement. \citet{hamilton2022covid} fine-tune a pre-2020 GPT-2 on CBC News articles, prompting it with real pandemic headlines to generate a counterfactual corpus of what COVID-naive coverage might have looked like, compared against actual reporting to surface human narrative framing choices. \citet{liu2026retell} apply reinforcement learning using \citet{todorov19712}'s narrative equilibrium principles as a theory-derived reward signal generated by an LLM-as-judge, with GRPO and LoRA. If narrative attributes can be formalised as reward signals in this way, RL post-training may offer a route to theory-informed story generation that bypasses the need for large-scale human annotations. All three are constrained in scope: the first two target specific structural transformations rather than open-ended generation, and \citet{liu2026retell}'s approach, while generalisable in principle, is demonstrated only on short counterfactual stories.

\subsection{\redtext{In-Context Learning}}

In-context learning techniques have been prominently explored and are highly effective in decoder-only models compared to unsupervised and supervised learning for narrative generation and understanding tasks (see \redtext{T}able \ref{tab:focus_table}). Prompting-only (or in-context learning only) approaches have been selected for a variety of narrative tasks. Various degrees of success have been reported using zero-shot or one-shot prompting. It seems that prompting alone is at times sufficient in applying narrative theories for different downstream tasks \citep{piper-etal-2024-social,piper-bagga-2024-using,shen_heart-felt_2024,tian-etal-2024-large-language,sui_confabulation_2024,chun_aistorysimilarity_2024,marino2024integrating}. There is a trend of simpler prompting strategies prevailing over complex ones due to similar performance and better efficiency \citep{zhou-etal-2024-large, chhun2024language}.

\cite{bartalesi2024using} found that small changes in prompt word choices can affect model performance. Furthermore, the attempt to include special characters as markers did not significantly increase performance. As LLMs can be sensitive towards minor variations in the prompts \citep{mire-etal-2024-empirical,bartalesi2024using}, researchers such as \citep{mire-etal-2024-empirical} use five paraphrases of the original question, and then use the per-text majority vote among the five independent output predictions as the final label. Select studies specify inference parameters such as temperature, top-p and frequency penalty \citep{halperin2024artificial,zhou-etal-2024-large,hobson-etal-2024-story,chun_aistorysimilarity_2024} which helps with replicability. The output variability from giving LLMs different instruction wordings and parameters is welcomed by some, as there are claims that the behaviour can decently mimic multiple human annotators \citep{troiano_clause-atlas_2024,harel-canada-etal-2024-measuring,chhun2024language,zhou-etal-2024-large}.


\redtext{\subsubsection{Zero-Shot and Few-Shot}}

By including narrative theory concepts as prompt content, zero-shot and few-shot tests whether LLMs can apply narratological frameworks without training examples. \citet{gervas2024tagging} evaluate whether LLMs can annotate folktale synopses with Proppian character functions directly via zero-shot and few-shot prompts, finding that models apply theory-derived labels with varying accuracy. \citet{piper_narradetect_2025}'s NarraDetect uses zero-shot prompting with theoretically defined criteria--agency, event sequencing, and world building from \citet{herman2009basic}--to classify narrative versus non-narrative passages across diverse text types. The relative success of zero-shot theory-driven annotation suggests that LLMs have absorbed some narrative frameworks during pre-training; the remaining failures reveal the limits of that absorption, particularly for theories like Proppian morphology, where LLMs can describe the framework in general terms but confuse character types with narrative functions and omit key functions when actually applying the taxonomy to stories \citep{gervas2024tagging}.

When zero-shot prompting is often chosen \citep{troiano_clause-atlas_2024,halperin2024artificial,zhou-etal-2024-large,piper-wu-2025-evaluating,tian-etal-2024-large-language}, a significant reason is cost consideration \citep{zhou-etal-2024-large,bartalesi2024using}. \cite{troiano_clause-atlas_2024} compare zero-shot and few-shot prompts, they observe that the outputs of gpt-3.5-turbo from different prompts converge more often than not and this variability is acceptable by comparing it to human annotations. They compare using Fleiss’ scores. They conclude that the diversity of outputs from different prompting strategies can represent the variations between human annotators. \cite{piper-wu-2025-evaluating} use a zero-shot prompting framework and compare 4 different LLMs: GPT-4o, Llama3:8B, Llama3.1:8B and Gemma2:9B. They find that introducing pre-processing of passages, such as summarisation, resulted in poorer model responses. The study found that pre-processing or intermediate steps were not needed as a zero-shot prompting approach performed on par with human annotations at topic labelling. 

In comparison, \cite{sun-etal-2024-event} develop a few-shot prompting to extract causal events from distinct story events. They find that their technique can generalise to different LLMs such as Yi-34B-chat, Llama-2-13B-chat. It also outperformed supervised learning methods on other language models such as GPT2-large and T5 large. To improve performance, \cite{marino2024integrating} prioritised a detailed prompt over one that optimised cost using GPT-4-turbo. The prompt includes a system prompt for context (which includes a role prompt) and output format, and a user prompt supplying necessary documents. It was crafted using strategies from OpenAI's prompt engineering guide\footnote{\url{https://platform.openai.com/docs/guides/prompt-engineering}}. 

Role-prompting is popular strategy chosen by several methodologies \citep{piper-bagga-2024-using,troiano_clause-atlas_2024,chun_aistorysimilarity_2024,marino2024integrating,bae-kim-2024-collective}. \cite{piper-bagga-2024-using} use a zero-shot prompting framework that consists of: role prompt, framing question, ordinal scale, narrative feature, and individual passage to quantitatively classify the degree of presence of narrative discourse features \citep{genette1980narrative,herman2009basic}. \cite{harel-canada-etal-2024-measuring} introduce a Mixture-Of-Personas (MoP) prompting strategy which builds on prior research demonstrating that large language model (LLM) performance can be improved by instructing them to adopt the persona of a domain expert story writer. Each persona is designed to focus on a specific component of the Psychological Depth Scale, such as an AI specialising in empathy or one that assesses narrative complexity. By averaging the judgements of these varied personas, the evaluation gains a diversity of opinion that better reflects multiple human annotators. The study also uses the plan and write approach similar to chain of thought. \redtext{However, the effective of role-prompts have also been called into question across broad domains \citep{zheng-etal-2024-helpful}.}


\subsubsection{\redtext{Chain of Thought}}

Chain-of-thought and multi-step prompting strategies decompose complex narrative reasoning into intermediate steps, making implicit interpretive work explicit as part of the prompt sequence. This is particularly relevant for narrative tasks, where properties such as a story's moral, causal structure, or quality are not directly recoverable from surface text but must be inferred through intermediate judgements. Across the surveyed work, this decomposition takes two forms: sequential pipelines where each prompt builds on the output of the last and iterative refinement across multiple passes.

\citet{hobson-etal-2024-story} apply sequential decomposition to story moral extraction: rather than prompting for a moral directly, they issue a sequence of prompts that progressively build narrative context--summarising the story, identifying the protagonist and antagonist, extracting the central topic and valence--before finally prompting for the moral. Each step provides structured context for the next. 


\citet{chakrabarty2024art} use iterative prompting for story generation--expanding the story across multiple passes to reach a target length, motivated by the observation that LLMs produce outputs 20--50\% shorter than intended. For evaluation, they apply chain-of-thought prompting to answer binary criteria questions assessing creative writing dimensions, though none of the LLM assessments correlated positively with expert judgements and few-shot prompting did not recover this.

\subsection{\redtext{Compound Systems}}
\label{subsec:compound}
\subsubsection{\redtext{Graphs and RAGs}}

Graph-based and retrieval-augmented (RAG) systems augment LLMs with structured or retrieved knowledge that lies outside or beyond the immediate prompt context across understanding and generation tasks, and both predominantly address fabula and discourse levels.

Graphs are used across both story generation and understanding tasks, where structured representations of events, causal relationships, and character networks serve as intermediary representations between system input and output. These systems leverage graphs as scaffolding for generation \citep{yoo2024leveraging,song2024conflict} and as outputs of understanding \citep{hamilton-etal-2025-city,das-etal-2024-media,heddaya-etal-2024-causal,sun-etal-2024-event}.

In generation contexts, graphs represent the fabula level by encoding causal chains, character goals, and event sequences that guide LLM output. \citet{yoo2024leveraging}'s Story Plan Graph system constructs a branching graph of possible narrative events, using LLMs to select and traverse a single coherent path before outputting the story in textual prose. \citet{ghaffari-hokamp-2025-narrative}'s Narrative Studio models narratives as dynamic causal event graphs, using Monte Carlo Tree Search alongside LLMs to explore and navigate possible story continuations, grounding narrative progression in \citet{todorov19712}'s theory of causal state transformations of the narrative equilibrium. \citet{song2024conflict}'s CNGCI adopts a neuro-symbolic approach, pairing a commonsense knowledge graph \citep{hwang2021comet} with a transformer to infer protagonist goals before introducing \citet{freytag1894technik}-derived causal obstacles, making the graph a prerequisite for structurally grounded conflict generation.

In understanding contexts, graphs are the target output rather than the generative scaffold. \citet{hamilton-etal-2025-city} use LLMs to extract social networks from multilingual fiction and non-fiction narratives, constructing character graphs to conduct large-scale distant reading analysis informed by \citet{moretti2011network}'s network theory. \citet{das-etal-2024-media} extract event-centric narrative graphs from news media to identify high-level narrative frames, while \citet{heddaya-etal-2024-causal} construct causal micro-narrative graphs from contemporary and historical news corpora, also draws on \citet{todorov19712}'s theory of narrative equilibrium to define the relational structure between events. \citet{sun-etal-2024-event} use graphs as an evaluation intermediary: prompting an LLM to generate a causal event graph of a story before producing an overall quality score, finding incremental improvements over non-graph baselines in correlation with human ratings.

RAG is used in two ways: external corpus retrieval to enrich understanding tasks, and internal story-memory retrieval to maintain coherence in generation. In understanding, \citet{wilmot2021memory} augment a transformer language model with an external knowledge base derived from Wikiplots--a corpus of story plot summaries--via RAG, alongside a memory mechanism that handles the scale of long-form literary texts. Their system targets event salience detection grounded in \citet{barthes1975introduction} Cardinal Functions, using chapter-aligned summaries from the Shmoop corpus as annotations. The combination of external knowledge retrieval and memory augmentation outperforms both non-RAG and non-memory baselines, demonstrating that narrative understanding tasks requiring long-range contextual reasoning benefit from explicit external knowledge access. \citet{liu2025raemollm}'s RAEmoLLM takes a different approach, building a retrieval database of affective embeddings from labelled source-domain data and retrieving the top-K emotionally similar examples for few-shot ICL on cross-domain misinformation detection. Referenced to \citet{keen2006theory}'s theory of narrative empathy, the system exploits the observation that fake news, rumours, and conspiracy theories carry distinct emotional signatures--retrieving emotionally similar labelled examples bridges domain gaps without fine-tuning, achieving improvements of up to 31\% over zero-shot baselines.

In generation, \citet{wang2025generating}'s DOME uses a temporal knowledge graph (KG)--a structured database that stores story entities and their relationships as a network of connected nodes--to store and retrieve previously generated story content for long-form story generation. As each section of the story is written, relevant prior content is retrieved via LLM-filtered KG queries, keeping historical information concise and reducing contextual conflicts--a problem DOME reports reducing by 87.61\% over non-memory baselines. Unlike external-corpus RAG, DOME's retrieval operates entirely within the developing story world, making the KG a form of structured narrative memory rather than an external knowledge source. \citet{chen2026storybox}'s StoryBox follows a similar logic at larger scale, using a dual-memory retrieval system combining keyword and dense vector search over previously generated events.

Across both graph-based and RAG systems, structured knowledge compensates for the limitations of direct LLM generation or understanding at scale. For understanding tasks, external retrieval brings in knowledge the model cannot reliably generate from parameters alone; for generation tasks, internal retrieval maintains narrative consistency across lengths that exceed context windows. Narrative representations across these systems is mostly at the fabula and discourse levels--event salience, causal structure, character networks--while narration and situatedness levels retrieval remains largely unexplored, except by \citet{liu2025raemollm}.


\subsubsection{\redtext{Multi-Agent Systems}}
\label{sec_mas}

\redtext{The most consistent theoretical pattern across narrative Multi-Agent Systems (MAS) is the application of the fabula/discourse distinction as a two-stage pipeline: one component simulates or plans story-world events, another generates the discourse in textual prose. \citet{yu-etal-2025-multi} implement this using a Director Agent to coordinate character agents in a role-play step that generates fabula content, then a Rewrite Agent that transforms the resulting interaction history into finished story text aligned with a specified presentation order (discourse). \citet{ran-etal-2025-bookworld}'s BOOKWORLD follows a similar implementation: character agents with goals, states, and memories interact across a simulated world to produce event records, which a separate agent then rephrases into ``novel-style story" with control over narrative pace. \citet{chen2026storybox}'s StoryBox uses Character Agents to simulate plot events while also recording each agent's contextual metadata including emotions and motivations, before a single Storyteller Agent drafts the final narrative. Focusing more on discourse than fabula, \citet{lima2024pattern}'s PatternTeller uses a Pattern Specialist Agent to extract and apply narrative structures such as the Hero's Journey, a Storywriter Agent for prose generation, and a Plot Manager to coordinate between them. In all these systems, what classical narratology described as a theoretical distinction between story world and textual presentation has become a popular design for narrative MAS.}

\redtext{For contrast, non-theory-informed MAS approaches seem to focus on feedback and iterative quality control. \citet{bae-kim-2024-collective}'s CRITICS system deploys a leader agent coordinating diverse critic agents, each assigned specific evaluation criteria--logical flow, lexical diversity--who engage in multi-turn debate to review and revise generated text. \citet{pei2024swag}'s SWAG uses a generator-evaluator loop, scoring potential next narrative actions for relevance and coherence before committing to prose, preventing plot drift--where the story gradually loses its intended direction across successive generations--and self-contradiction common in auto-regressive generation. \citet{venkatraman2025collabstory}'s CollabStory takes a different angle, deploying multiple foundation models as turn-taking co-authors to simulate human collaborative writing dynamics and study AI authorship signatures. While these approaches are not narrative-theory informed, they show the MAS offers the potential to model stories beyond the fabula and discourse levels, and instead on the narration and situatedness level. We state this to show that there is potential for MAS to model those levels of narrative, which is currently unexplored in a theory-informed way.}

\section{Summary of Surveyed Research Outcomes}
\label{sec_outcomes}

We group survyed research outcomes into five strands: (i) methodological and performance gains; (ii) new data, annotations, and benchmarks; (iii) insights into LLM capabilities, biases, and cultural framing; (iv) domain-specific narrative findings; and (v) theoretical and conceptual developments.

\textbf{Targeted modelling choices can improve specific narrative tasks over unstructured generation baselines.} Applying narratological distinctions between fabula, discourse, and narration as structural principles--whether in structured multi-step generation approaches \citep{halperin2024artificial, chen-si-2024-reflections} or multi-agent pipelines \citep{yu-etal-2025-multi, ran-etal-2025-bookworld}--yields gains in narrative coherence, controllability, and task-specific objectives. At the same time, human-authored stories still consistently outperform LLM-generated ones on criteria including originality, surprise, and rhetorical complexity \citep{chakrabarty2024art, tian-etal-2024-large-language, bissell-etal-2025-theoretical}.

\textbf{Theory-informed narrative datasets and annotations have grown.} Contributions include: StorySeeker \citep{antoniak-etal-2024-people}, 502 Reddit posts annotated with story and event spans across 33 topic categories; NarraDetect \citep{piper-bagga-2025-narradetect}, scalar narrativity ratings across 18 genres including memoirs, academic articles, and book reviews; HeartfeltNarratives \citep{shen_heart-felt_2024}, personal stories annotated with empathy-evoking narrative features; TUNA \citep{brei-etal-2025-classifying}, a multi-genre dataset annotated for three levels of narrator unreliability; PolyNarrative \citep{nikolaidis-etal-2025-polynarrative}, a multilingual, multi-domain news narrative dataset; and CR4-NarrEmote \citep{piper2025cr4}, an open-vocabulary dataset of over 200,000 citizen science annotations of narrative emotions across thousands of literary passages. Despite this growth, standardised benchmarks remain severely limited: \citet{hamilton-etal-2026-narrabench} survey 78 existing benchmarks and find that only 27\% of narrative tasks are well-captured, with narrative events, style, perspective, and revelation nearly absent from current evaluations.

\textbf{Findings produce a clearer view of LLM capabilities and biases.} Cross-cultural studies show that name cues trigger stereotypes, with Western names linked to wealth and Arab names to modesty \citep{naous-etal-2024-beer}. \citet{rooein2025biased} find that appearance-related attributes in LLM-generated children's stories increase by over 55\% for girl protagonists, and that stories featuring non-Western children disproportionately emphasise cultural heritage and family themes. \citet{ghosal2026ai} find that ChatGPT and Gemini systematically default to a hero's journey arc privileging economic assimilation and positive affect, while marginalising trauma, irregular migration status, and non-white immigrant experiences in narratives about noncitizenship. LLM-generated stories are also embedded with canonical literary structures such as the Pygmalion myth \citep{beguvs2024experimental}, cluster at uniformly high positive valence and low arousal relative to human writing \citep{tian-etal-2024-large-language}, and may compress the diversity of authorial voices through instruction tuning \citep{hamilton-2024-detecting}. LLMs nonetheless perform well on many narrative understanding tasks: extracting latent moral framing \citep{zhou-etal-2024-large}, character social networks from multilingual fiction and non-fiction at scale \citep{hamilton-etal-2025-city}, and discourse-level features such as tense, mood, and voice \citep{piper-bagga-2024-using}.

\textbf{Context-specific insights are enabled by scalable narrative analysis.} \citet{antoniak-etal-2024-people} find that storytelling rates vary widely across Reddit: highest in personal experience communities (r/tifu: 0.98 stories per post), prevalent in healthcare communities, and lowest in topic-focused ones such as r/Futurology. \citet{zhou-etal-2024-large} find that Chinese and North American climate reporting differs in moral messaging and emotional valence. \citet{hobson-etal-2024-story} find that moral schemas vary across cultural and generic contexts, including folktales, novels, news, and social media. \citet{mitran2025probing} apply Moral Foundations Theory \citep{haidt2004intuitive} to 2,697 folktales from 55 countries, finding broad cross-cultural consistency in moral foundations but more balanced positive and negative moral content than prior work suggested. \citet{ma2026multi} find that across 830 Chinese therapeutic texts, narrative organisation predicts mental health severity better than vocabulary, with depression and anxiety showing distinct structural signatures such as temporal disorientation.

\textbf{LLM methods can contribute to theoretical development and validation}, covered in detail in Section~\ref{subsec:theoretical_contributions}. Surveyed work contributes through synthesis of existing frameworks, application of theories into measurable taxonomies, and empirical validation of theoretical claims. On the synthesis side, \citet{piper-bagga-2024-using} combine Genette's and Herman's frameworks into fifteen measurable discourse features; \citet{shen_heart-felt_2024} merge Keen's narrative empathy theory with Van Krieken et al.'s sociolinguistic cues into the HEART taxonomy; and \citet{harel-canada-etal-2024-measuring} synthesise reader-response criticism and cognitive narrative theory into a Psychological Depth Scale. For empirical validation, \citet{hamilton-etal-2025-city} validate Moretti's network theory by finding that fictional social networks are significantly smaller and more tightly clustered than non-fictional ones; \citet{beguvs2024experimental} provide empirical grounding for the persistence of the Pygmalion cultural schema across both human and LLM storytelling; \bluetext{\citet{bissell-etal-2025-theoretical} adapt \citet{sternberg1990telling}'s theory of surprise into six annotation criteria, confirming that human-authored endings are more surprising and more preferred than LLM-authored ones; and \citet{liu2026retell} encode Todorov's narrative equilibrium directly as a reward signal for reinforcement learning, finding that the presence of narrative structural stages correlates with human narrative preferences.}

\section{Challenges and Future Directions}\label{sec_future}

\redtext{\citet{piper-2023-computational} outlined eight challenge areas for computational narrative understanding, arranged in increasing order of generality and complexity: dataset creation, narrative element detection, multilingual modelling, multimodal modelling (outside the scope of our survey), narrative discourse detection, narrative time modelling, narrative schemas and taxonomies, and collective stories and social behaviour. Our survey of 68 papers published between 2021 and 2026 suggests that the LLM era has made meaningful but uneven progress across this list. The earlier challenges--dataset creation, narrative element detection, and narrative discourse detection--are represented in current research. The latter challenges, however, remain largely unaddressed: multilingual modelling, narrative time modelling, and collective stories and social behaviour. The following sections discuss how these gaps are reflected in the surveyed research and suggest future directions.}

\subsection{Challenges}
    
    It is already well recognised that progress in computational narrative understanding and generation is constrained by the lack of standardised datasets, evaluation metrics, and theoretical frameworks, which complicates model assessment and system development \citep{wang2023open, alhussain2021automatic, alabdulkarim-etal-2021-automatic}. 

    \textbf{LLMs have advanced narrative generation substantially in fluency, coherence, and contextual consistency relative to earlier methods \citep{wang2023open}, yet significant shortcomings persist across multiple dimensions.} Generated stories are often structurally flat, clustering at uniformly high positive valence and low arousal relative to human writing \citep{tian-etal-2024-large-language}, and fall short of professional authors on originality, surprise, and rhetorical complexity \citep{chakrabarty2024art, bissell-etal-2025-theoretical}. \citet{brei2026casper} finds that LLM-written characters systematically display stereotypical portrayals, transparent motivations, and neatly completed arcs dominate regardless of model size, though model family and genre both influence the degree of character variety produced. Even for summarisation, faithfulness remains a problem: \citet{subbiah2024reading} find that GPT-4, Claude-2.1, and LLaMA-2-70B all make faithfulness mistakes in over 50\% of short story summaries when evaluated directly by the stories' authors. \citet{ahuja2025finding}'s FLAWEDFICTIONS benchmark for evaluating narrative reasoning through plot hole detection shows that state-of-the-art LLMs struggle regardless of reasoning effort, with performance degrading sharply as story length increases; LLM-based story summarisation and generation introduce over 50\% and 100\% more plot holes respectively than human-written originals, quantifying a persistent gap in narrative consistency. Story understanding tasks, by contrast, show comparatively stronger results: LLMs can label narrative topics on par with human annotators in zero-shot settings \citep{piper-wu-2025-evaluating}, achieve strong performance on character interaction classification after fine-tuning \citep{piper-etal-2024-social}, and segment narrative events at rates similar to human annotators \citep{michelmann_large_2025}.
    
    \textbf{Consistent evaluation of story generation remains problematic, with conflicting claims about LLM performance}--some studies report underperformance relative to human writers \citep{chakrabarty2024art, tian-etal-2024-large-language, beguvs2024experimental, bissell-etal-2025-theoretical}, while others find LLMs outperform earlier story generation models and compete with human authors on fluency and coherence \citep{xie-etal-2023-next}, or find that GPT-4-generated stories are statistically indistinguishable from highly-rated human-authored stories on reader-centred dimensions of authenticity, empathy, and emotional engagement \citep{harel-canada-etal-2024-measuring}. These discrepancies reflect inconsistent metrics and task definitions \citep{bissell-etal-2025-theoretical}. However, this reflects the broader fact that there are no agreed upon standards of evaluation when it comes to the quality or aesthetic value of a literary narrative. 
    
    \textbf{Many challenges faced in this field are replications of human problems}. If the broader narrative studies landscape cannot reach a consensus on shared knowledge and terminology, then it is unsurprising that the lack of standardisation continues to remain a challenge in NLP. While we have no solution or ability to imagine what a standardised knowledge of narrative might look like, we believe that increasing awareness of the narrative studies landscape would help researchers in making conscious and intentional efforts and collaborations. 
    
   \textbf{Cultural and demographic bias is a persistent challenge in LLM narrative research.} Models reproduce stereotypes from training data: name cues trigger cultural associations linking Western names to wealth and Arab names to modesty \citep{naous-etal-2024-beer}; appearance attributes increase by over 55\% for girl protagonists \citep{rooein2025biased}; and narratives about non-citizenship default to hero's journey arcs that marginalise non-white and irregular migration experiences \citep{ghosal2026ai}. Importantly, human crowd-workers show similar biases in the stories they produce \citep{beguvs2024experimental}, suggesting the problem lies in training data and annotation practice as much as model architecture.

    \textbf{Dataset challenges including multilingual corpora \citep{piper2022conlit}, human-annotated datasets \citep{shokri-etal-2025-finding}, and labelled long-form narratives \citep{yamshchikov-tikhonov-2023-wrong}}. Existing datasets such as FairytaleQA \citep{xu-etal-2022-fantastic} and corpora from Project Gutenberg are English-only, as are studies using Hollywood scripts \citep{chun_aistorysimilarity_2024,tian-etal-2024-large-language}. Theoretical work also centres on Western narrative traditions \citep{piper2021narrative}, with limited attention to cultural diversity \citep{sui_confabulation_2024}. Recent efforts have adapted Western frameworks for non-English contexts, such as developing systematic guidelines to apply Labovian structural analysis to Japanese oral narratives \citep{watahiki2026developing}. The cross-cultural study of narrative theories remains under-explored \citep{piper-bagga-2024-using,shen_heart-felt_2024,michelmann_large_2025}.

\subsection{Future Directions}

    We believe that there is much more to narrative studies than ``what is a narrative text and what makes it good?'', which is a difficult question to answer. Instead, meaningful outcomes may arise when we move away from storytelling as an all-encompassing idea/entity as the object of study and instead focus on more tangible specifics. As such, we suggest the following possible approaches for future research:



    \begin{description}[itemsep=0.5em, labelwidth=!, leftmargin=0pt]

        \item[\textbf{Continual engagement with narrative studies beyond narratology:}] While current research takes inspiration mainly from classical narratology and focuses on literary narratives such as fairytales, we believe LLMs' advantages of scale, generalisation and distant reading capabilities make them valuable tools for broader narrative studies fields. Future research could shift focus to existing areas of research, such as history's narrativisation and emplotment of events \redtext{or the} focus of International Relations on the creation and circulation of `strategic narratives' in geopolitical soft power struggles between nation states, and developing methods for harvesting or producing valuable data. The object of study in this case are the social, historical, and cultural phenomenons of narrative.
        \item[\textbf{Validate specific narrative theories:}] Find a narrative theory that makes a claim about narrative. \redtext{e.g.,} \cite{keen2006theory}'s claim that fiction offers readers unique empathetic engagements compared to other genres. Then design a system that tests the validity of that theory and its assumptions. 
        \item[\textbf{Develop specific metrics:}] Instead of ambitiously aiming for an all-encompassing benchmark that measures a narrative as a whole, focus on developing metrics that measure only a single aspect of narrative. Perhaps the capacity to track certain stylistic features of narrative both within a work and across a larger corpus.
        \item[\textbf{Investigate the narrative impact on pre-training:}] While ongoing research focus mainly on in-context learning and post-training, it cannot be ignored that narrative texts make up large portions of LLM pre-training. What effect does pre-training on more or less narrative texts have on an LLM?
        \item[\textbf{LLM interpretability through narrative studies:}] Narratives have been hypothesised to be a fundamental method for knowledge representation in AI systems \citep{szilas2015towards}. Perhaps current LLMs already use narratives as their method for knowledge representation. It can be argued that LLM outputs, intersect with narrative studies as they inherently reference the textual narratives seen during the pre-training process.
        \item[\textbf{Investigate post-training for story generation:}] Section \ref{finetune} showed that research on fine-tuning LLMs for story generation is close to non-existent. Most LLMs used for research are `instruction-tuned' variants. If we are to believe the narrative-first hypothesis \citep{turner1996literary,mcbride2014storytelling,ferretti_narrative_language_2024}, that language emerged to facilitate storytelling, not instruction following, can there possibly be a `narration-tuned' LLM?

    \end{description}
        
\section{Conclusion}\label{sec_conclusion}
    To the best of our knowledge, our survey is the first that takes an NLP-oriented view to automatic story generation and understanding tasks with a focus on broader interdisciplinary narrative studies, and analyses the theoretical influences of various narrative theories drawn by ongoing research. \redtext{Our survey provides a systematic mapping between narratological frameworks and state-of-the-art LLM pipelines that prior work has not attempted.} Just as a story has a beginning, middle and end \citep{aristotle_poetics_penguin}, our survey began by introducing key terminology from narrative studies to introduce a much-needed taxonomy that contextualises the trends in the field; we then explored the current climate of relationships between narrative theories, datasets, tasks and LLM methods; finally, we end the survey by discussing challenges and future directions.

    While many challenges are common to most artificial intelligence fields (more data, more variety, more labels), through this survey we hope to emphasise that significant gaps exist in the engagement with concepts from narrative studies scholarship. Instead of hoping for standardised benchmarks or metrics for narrative that may never arrive, we believe it is advantageous to engage with specific sub-fields of narrative studies through intentional approaches.

    Specifically, by surveying 68 papers published between 2021 and 2026, mapping their theoretical foundations, datasets, narrative levels, LLM methods, and research outcomes, we find the following insights:

\begin{itemize}

    \item \textbf{\redtext{Literary theories are travelling beyond their original 
    contexts.}} The majority of surveyed works draw theories from lineages relating to classical narratology (Table \ref{sec_narrative_theories}), yet many apply it to non-literary texts such as news and social media at scale. This cross-domain application is a novel experimentation involving theories originally designed for literature.

    \item \textbf{\redtext{Generation lags behind understanding in theoretical and technical 
    application.}} We find more examples of theory-informed understanding methods than generation methods. Understanding methods also tend to draw from more diverse sources of narrative theory while generation methods apply mostly classical or structuralist narrative theories. Story generation tasks almost exclusively rely on prompt-engineering or compound systems, while post-training is applied consistently and effectively for understanding, with no established approach for open-ended, theory-informed post-training for story generation.

    \item \textbf{\redtext{Some narrative levels remain under-explored.}} 
    Narration and situatedness are addressed far less than fabula and discourse, 
    particularly in generation tasks, and existing benchmarks capture only a 
    fraction of the narrative phenomena that matter. Expanding research beyond \textit{what} a story contains and \textit{how} it is told, to \textit{who} is telling it and \textit{for whom} would bring NLP closer to the full scope of what narrative studies considers important.

    \item \textbf{\redtext{Computational methods can contribute back to theory.}} A 
    meaningful proportion of surveyed papers produce synthesis, taxonomy, or 
    validation outcomes, showing that large-scale empirical analysis can validate, 
    refine, or challenge theoretical claims in ways that close reading alone 
    cannot. This bidirectional potential extends to pre-training, which remains 
    largely uninvestigated despite suggestive evidence that narrative structure is 
    actively reinforced by the objectives and architectures through which models 
    learn. With billions of parameters trained on more narrative text than any 
    other artefact in human history, the question of \textit{what kind of 
    narrative artefact is an LLM?} may unlock new understanding of both model 
    behaviour and narrative theory itself. Computational findings can generate new hypotheses, expose the limits of existing frameworks, and open lines of inquiry that neither field could pursue alone.

\end{itemize}

    \section*{Acknowledgements}
        We would like to thank Sebastian Sequoiah-Grayson for providing feedback and support during the drafting process.
    
    \section*{Funding}
        This research was supported by the Commonwealth \bluetext{of Australia} through an Australian Government Research Training Program Scholarship [DOI: https://doi.org/10.82133/C42F-K220].

    \section*{Declaration of Generative AI Assistance}
        Parts of the manuscript were drafted or rephrased for grammatical accuracy with the assistance of Claude, Microsoft Copilot and Anara. The authors reviewed and revised all AI-generated text and take full responsibility for the content. The two figures were made using Miro.

    
\bibliography{david}

\begin{appendices}
\end{appendices}

\end{document}